\documentclass{article}

\usepackage[preprint]{neurips_2026}

\usepackage[utf8]{inputenc} 
\usepackage[T1]{fontenc}    
\usepackage{booktabs}       
\usepackage{amsfonts}       
\usepackage{nicefrac}       
\usepackage{microtype}      
\usepackage{xcolor}         

\usepackage{graphicx}
\usepackage[hyphens,spaces,obeyspaces]{url}
\usepackage[breaklinks]{hyperref}

\usepackage{amsmath}
\usepackage{amssymb}
\usepackage{mathtools}
\usepackage{amsthm}
\usepackage{macros}
\usepackage{subcaption}
\usepackage[textsize=tiny]{todonotes}
\usepackage{wrapfig}

\title{Simulus: Combining Improvements in Sample-Efficient World Model Agents}

\author{%
    Lior Cohen \\
    Technion \\
    \And
    Kaixin Wang \\
    Microsoft Research \\
    \And
    Bingyi Kang \\
    ByteDance Seed \\
    \And
    Uri Gadot \\
    Technion \\
    \And
    Shie Mannor \\
    Technion \\
}

\begin{document}

\maketitle

\begin{abstract}
World models (WMs) represent the frontier of sample-efficient reinforcement learning, but their complexity leaves many promising improvements unrealized due to the significant expertise and effort required to identify and integrate them. 
Inspired by Rainbow, which showed that individually known improvements to DQN complement each other and can be effectively combined, we take on this challenge and ask whether the same principle applies to world model agents.
We introduce Simulus, a modular token-based WM agent that integrates: (1) a flexible tokenization framework supporting arbitrary combinations of observation and action modalities; (2) intrinsic motivation for epistemic uncertainty reduction; (3) prioritized world model replay; and (4) regression-as-classification for reward and return prediction. 
Simulus achieves state-of-the-art sample efficiency for planning-free WMs across three diverse benchmarks: visual Atari 100K, continuous-control DMC Proprioception 500K, and symbolic Craftax-1M. 
Notably, intrinsic motivation proves beneficial even under the tight interaction budgets of sample-efficient RL, despite the risk of wasting scarce interactions on task-irrelevant experience. 
Ablation studies reveal that each component contributes individually, and their combination yields synergistic gains.
Our code and model weights are publicly available at
\url{https://github.com/leor-c/Simulus}.

\end{abstract}

\section{Introduction}


%
%
%

Sample efficiency refers to the ability of a reinforcement learning (RL) algorithm to learn effective policies using as few environment interactions as possible. 
In many real-world domains such as robotics, autonomous driving, and healthcare, this is particularly critical, as interactions are costly, slow, or constrained.
World model agents, methods that learn control entirely from simulated experience generated by a learned dynamics model, have emerged as a promising approach to improving sample efficiency \citep{ha2018worldmodels, hafner2023mastering, micheli2022transformers, wang2024efficientzero}.

Among world model agents, token-based world models (TBWMs) \citep{micheli2022transformers, micheli2024efficient, cohen2024improving} are a particularly promising direction because they cast environment dynamics as sequence modeling over discrete tokens. This formulation aligns naturally with the success of large-scale sequence models, where multi-token representations have proven effective for modeling complex data.
Evidently, most large scale world models \citep{agarwal2025cosmos, oasis2024, genie2} operate on multi-token observations, suggesting that such representations are advantageous at scale.
TBWMs offer a clear modular design, separating the optimization of its representation, dynamics, and control models. 
As modular systems, TBWMs are easier to scale, develop, study, and deploy, as individual modules can be treated independently without interfering and are easier to master through divide and conquer.
In addition, such separation leads to simpler optimization objectives and avoids interference between them (Appendix~\ref{sec:appendix-rssm-interference}).

\begin{figure}[t]
    \centering
    \includegraphics[width=\linewidth]{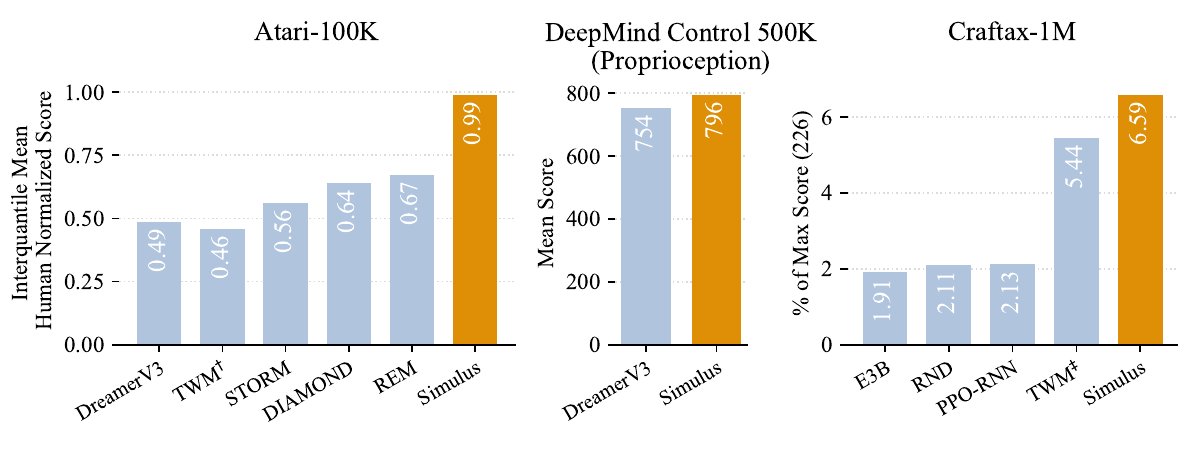}
    \caption{Results overview. \AlgName{} exhibits state-of-the-art sample-efficiency performance for planning-free methods across all three benchmarks. $\dagger$ \citep{robine2023transformer}, $\ddagger$ \citep{dedieu2025TWM2}.}
    \label{fig:first-page-results}
\end{figure}

Despite this promise, the potential of TBWMs remains largely unrealized. First, existing TBWMs are restricted to image observations and discrete actions, limiting both their adoption and broader applicability. While multi-modality tokenization approaches exist for large-scale offline settings \citep{reed2022a, schubert2023generalist, lu2023unifiedio, Lu_2024_CVPR}, these methods rely on large vocabularies (e.g., 33K tokens) which are inefficient for online, data-limited regimes. Whether substantially smaller vocabularies can preserve competitive performance remains an open question. Second, despite compelling results \citep{Schmidhuber2010, sekar20aPlan2Explore, kauvar2023curiousReplay}, intrinsic motivation and prioritized \emph{world model} replay are often overlooked in sample-efficient WM agents \citep{hafner2021dreamerv2, hafner2023mastering, zhang2024storm, cohen2024improving, alonso2024diffusion}, particularly in combination. We believe this is a consequence of the high cost of integrating and validating new components in tightly coupled systems. We conjecture that intrinsic motivation is further underused in sample-efficient settings, as it may steer the agent toward task-irrelevant regions and waste the limited interaction budget. Prioritized replay \citep{kauvar2023curiousReplay}, while promising, lacks robust empirical support and proved challenging to tune.
This situation is reminiscent of the landscape that preceded Rainbow \citep{hessel2018rainbow}: a set of individually validated improvements that had nonetheless never been systematically combined. 
Rainbow demonstrated the benefits of combining such complementary improvements, and became a landmark in the deep RL literature.
We ask whether the same principle applies to world model agents, and introduce \AlgName{}, a modular token-based WM agent that integrates: (1) a tokenization framework for handling arbitrary combinations of observation and action modalities, (2) intrinsic motivation for epistemic uncertainty reduction \citep{pmlr-v97-shyam19a, sekar20aPlan2Explore}, (3) prioritized world model replay~\citep{kauvar2023curiousReplay}, and (4) regression-as-classification (RaC) \citep{farebrother2024stop} with exponential binning \citep{hafner2023mastering} for reward and return prediction.

To evaluate the impact of the proposed components, we conducted extensive empirical evaluations across three diverse benchmarks, ranging from the visual Atari 100K \citep{Kaiser2020Model}, to the continuous proprioception tasks of the DeepMind Control Suite \citep{tunyasuvunakool2020DMC}, to Craftax \citep{matthews2024craftax}, which combines symbolic 2D grid maps with continuous state features.
\AlgName{} achieves state-of-the-art sample efficiency for planning-free WMs across all three. Ablation studies confirm the individual contribution of each component, and reveal considerable synergies when they are combined.
Summary of contributions:
\begin{itemize}
    \item We identify and integrate promising but overlooked components (intrinsic motivation, prioritized replay, and a hybrid RaC), showing across three diverse benchmarks that each contributes significantly, with particularly strong gains when combined.
    \item We establish TBWMs as versatile agents by adapting multi-modality tokenization to the sample-efficient regime, with \AlgName{} achieving state-of-the-art planning-free performance across visual, continuous, and multi-modal benchmarks.
    
    
    \item We show that world-model-induced exploration significantly improves performance even under tight interaction budgets, despite the risk of wasting scarce interactions on task-irrelevant experience.

    \item A strong starting point for TBWM research: building a competitive WM agent with all the integrated components requires significant expertise, engineering effort, and careful tuning. We open-source our code and model weights to facilitate future research.
    
\end{itemize}





\section{Method}
\label{sec:method}

\paragraph{Notations}
We consider the Partially Observable Markov Decision Process (POMDP) setting.
However, since in practice the agent has no knowledge about the hidden state space, consider the following state-agnostic formulation.
Let $\obsSet, \actionsSet$ be the sets of observations and actions,  respectively.
At every step $t$, the agent observes $\obs_{t} \in \obsSet$ and picks an action $\action_{t} \in \actionsSet$.
From the agent's perspective, the environment evolves according to $\obs_{t+1}, \reward_{t}, \doneSgnl_{t} \sim p(\obs_{t+1}, \reward_{t}, \doneSgnl_{t} | \obs_{\leq t}, \actionsVec_{\leq t})$, where $\reward_{t}, \doneSgnl_{t}$ are the observed reward and termination signals, respectively.
The process repeats until a positive termination signal $\doneSgnl_{t} \in \{0, 1\}$ is obtained.
The agent's objective is to maximize its expected return $\E [\sum_{t=0}^{\infty} \discountF^{t} \reward_{t+1}]$ where $\discountF \in [0, 1]$ is a discount factor.

For multi-modal observations, let $\obs_{t} = \{ \obs_{t}^{(i)} \}_{i=1}^{|\modalitySet|}$ where $\modalitySet$ is the set of environment modalities and $\obs_{t}^{(i)}$ denotes the features of modality $\modalitySet_{i}$.

\paragraph{Overview}
\AlgName{} builds on REM \citep{cohen2024improving}.
The agent comprises a representation module $\Tokenizer$, a world model $\WM$, a controller $\Controller$, and a replay buffer.
To facilitate a modular design, following REM, each module is optimized separately.
The training process of the agent involves a repeated cycle of four steps: data collection, representation learning ($\Tokenizer$), world model learning ($\WM$), and control learning in imagination ($\Controller$).

\subsection{The Representation Module $\Tokenizer$}
$\Tokenizer$ is responsible for encoding and decoding raw observations and actions.
It is a modular tokenization system with encoder-decoder pairs for different input modalities. 
Encoders produce fixed-length token sequences, creating a common interface that enables combining tokens from various sources into a unified representation.
After embedding, these token sequences are concatenated into a single representation, as described in Section \ref{sec:wm-embedding}.
Note that encoder-decoder pairs need not be learning-based methods; however, when learned, they are optimized independently.
This design enables $\Tokenizer$ to deal with any combination of input modalities, provided the respective encoder-decoder pairs.

\paragraph{Tokenization}
$\Tokenizer$ transforms raw observations $\obs$ to sets of fixed-length integer token sequences $\tokens = \{ \tokens^{(i)} \}_{i=1}^{|\modalitySet|}$ by applying the encoder of each modality $\tokens^{(i)} = \encoder_{i}(\obs^{(i)})$.
Actions $\action$ are tokenized using the encoder-decoder pair of the related modality to produce $\actionTokens$. 
The respective decoders reconstruct observations from their tokens: $\hat{\obs}^{(i)} = \decoder_i(\tokens^{(i)})$. 

\AlgName{} natively supports four modalities: images, continuous vectors, categorical variables, and image-like multi-channel grids of categorical variables, referred to as "2D categoricals".
More formally, 2D categoricals are elements of $([k_1]\times[k_2]\times\ldots\times [k_C])^{m \times n}$ where $k_1, \ldots , k_C$ are per channel vocabulary sizes, $C$ is the number of channels, $m, n$ are spatial dimensions, and $[k] = \{ 1,\ldots, k \}$.
Further details on per-modality tokenization are available in Appendix \ref{sec:appendix-cts-vectors-tokenization}.


\subsection{The World Model $\WM$} 
\label{sec:method-wm}
The purpose of $\WM$ is to learn a model of the environment's dynamics.
Concretely, given trajectory segments $\tokensTraj_{t} = \tokens_{1}, \actionTokens_{1}, \ldots, \tokens_{t}, \actionTokens_{t}$ in token representation, $\WM$ models the distributions of the next observation and termination signal, and the expected reward:
\begin{align}
    \text{Transition:}\quad & p_{\theta}(\hat{\tokens}_{t+1} \vert \tokensTraj_{t} ), \\
    \text{Reward:}\quad & \hat{\reward}_{t} = \hat{\reward}_{\theta} ( \tokensTraj_{t} ), \\
    \text{Termination:}\quad & p_{\theta}(\hat{\doneSgnl}_{t} \vert \tokensTraj_{t}),
\end{align}
where $\theta$ is the parameters vector of $\WM$ and $\hat{\reward}_{\theta}(\tokensTraj_{t})$ is an estimator of $\E_{\reward_{t} \sim p(\reward_{t} \vert \tokensTraj_{t} )} [ \reward_{t} ]$.

\paragraph{Architecture} $\WM$ comprises a sequence model $\seqModel$ and multiple heads for the prediction of tokens of different observation modalities, rewards, termination signals, and for the estimation of model uncertainty.
Concretely, $\seqModel$ is a retentive network (RetNet) \citep{sun2023retentive} augmented with a parallel observation prediction (POP) \citep{cohen2024improving} mechanism.
All heads are implemented as multilayer perceptrons (MLP) with a single hidden layer.
We defer the details about these architectures to Appendix \ref{sec:wm-additional-details}.

\paragraph{Embedding}
\label{sec:wm-embedding}
$\WM$ translates token trajectories $\tokensTraj$ into sequences of $\retnetDmodel$-dimensional embeddings $\tknEmbs$ using a set of embedding (look-up) tables.
By design, each modality is associated with a separate table.
In cases where an embedding table is not provided by the appropriate encoder-decoder pair, $\WM$ and $\Controller$ learn dedicated tables separately and independently.
As embeddings sequences are composed hierarchically, we use the following hierarchical notation:
\begin{align*}
    \text{Observation-action block:}\quad & \tknEmbs_{t} = (\obsBlock_{t}, \actBlock_{t}), \\ 
    \text{Observation block:}\quad &  \obsBlock_{t} = (\obsModalityBlock{1}_{t}, \ldots, \obsModalityBlock{|\modalitySet|}_{t}) ,
\end{align*}
where $\tokensPerObs_{i}$ denotes the number of embedding vectors in $\tknEmbs^{(i)}_{t}$.
Similarly, $\tokensPerObs=\sum_{i=1}^{|\modalitySet|}\tokensPerObs_{i}$.
To combine latents of each $\tokens_{t}$, $\Tokenizer$ concatenates their token sequences along the temporal axis based on a predefined modality order.
We defer the full details on the embedding process to Appendix \ref{sec:wm-additional-details}.

\vspace{-25pt}
\begin{wrapfigure}{r}{0.52\linewidth}
    \hspace{0.1cm}
    \centering
    \begin{subfigure}[t]{0.45\linewidth}  
        \centering
        \includegraphics[width=\linewidth]{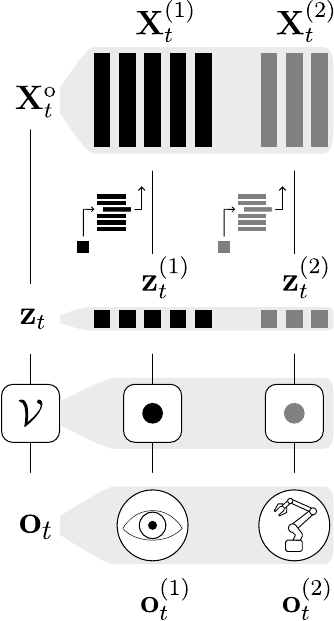}
        \caption{Observation tokenization and embedding.}
        \label{fig:tokenization-and-emb}
    \end{subfigure}
    \hfill
    \begin{subfigure}[t]{0.45\linewidth}  
        \centering
        \includegraphics[width=\linewidth]{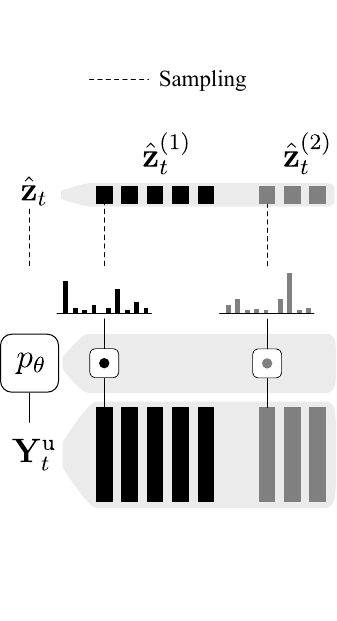}
        \caption{Prediction of observation tokens.}
        \label{fig:obs-pred}
    \end{subfigure}
    \caption{An illustration of the independent processing of modalities for an observation with two modalities.}
    \label{fig:tokenization}
\end{wrapfigure}
\vspace{1cm}
\paragraph{Sequence Modeling}
Given a sequence of observation-action blocks $\tknEmbs = \tknEmbs_{1}, \ldots, \tknEmbs_{t}$, the matching outputs $\retnetOutput_{1}, \ldots, \retnetOutput_{t}$ are computed auto-regressively as follows:
\begin{equation*}
    (\retnetState_{t}, \retnetOutput_{t}) = \seqModel(\retnetState_{t-1}, \tknEmbs_{t}),
\end{equation*}
where $\retnetState_{t}$ is a recurrent state that summarizes $\tknEmbs_{\leq t}$ and $\retnetState_{0}=0$.
However, the output $\retnetOutput^{\text{u}}_{t+1}$, from which $\hat{\tokens}_{t+1}$ is predicted, is computed using the POP mechanism via another call as
\begin{equation*}
    (\cdot, \retnetOutput_{t+1}^{\text{u}}) = \seqModel(\retnetState_{t}, \tknEmbs^{\text{u}}),
\end{equation*}
where $\tknEmbs^{\text{u}} \in \mathbb{R}^{\tokensPerObs \times \retnetDmodel}$ is a learned embedding sequence.
Intuitively, $\tknEmbs^{\text{u}}$ acts as a learned prior, enabling the parallel generation of multiple tokens into the future.


To model $p_{\theta}(\hat{\tokens}_{t+1} | \retnetOutput_{t+1}^{\text{u}})$, the distributions $p_{\theta}(\hat{\token} | \retnetOutputVec)$ of each token $\hat{\token}$ of $\hat{\tokens}^{(i)}_{t+1}$ of each modality $\modalitySet_i$ are modeled using modality-specific prediction heads implemented as MLPs with a single hidden layer and an output size equal to the vocabulary size of $\encoder_i$ (Figure \ref{fig:obs-pred}).
For 2D categoricals, $C$ heads are used to predict the $C$ tokens from each $\retnetOutputVec$.

\begin{figure*}[t]
    \centering
    \includegraphics[width=0.99\linewidth]{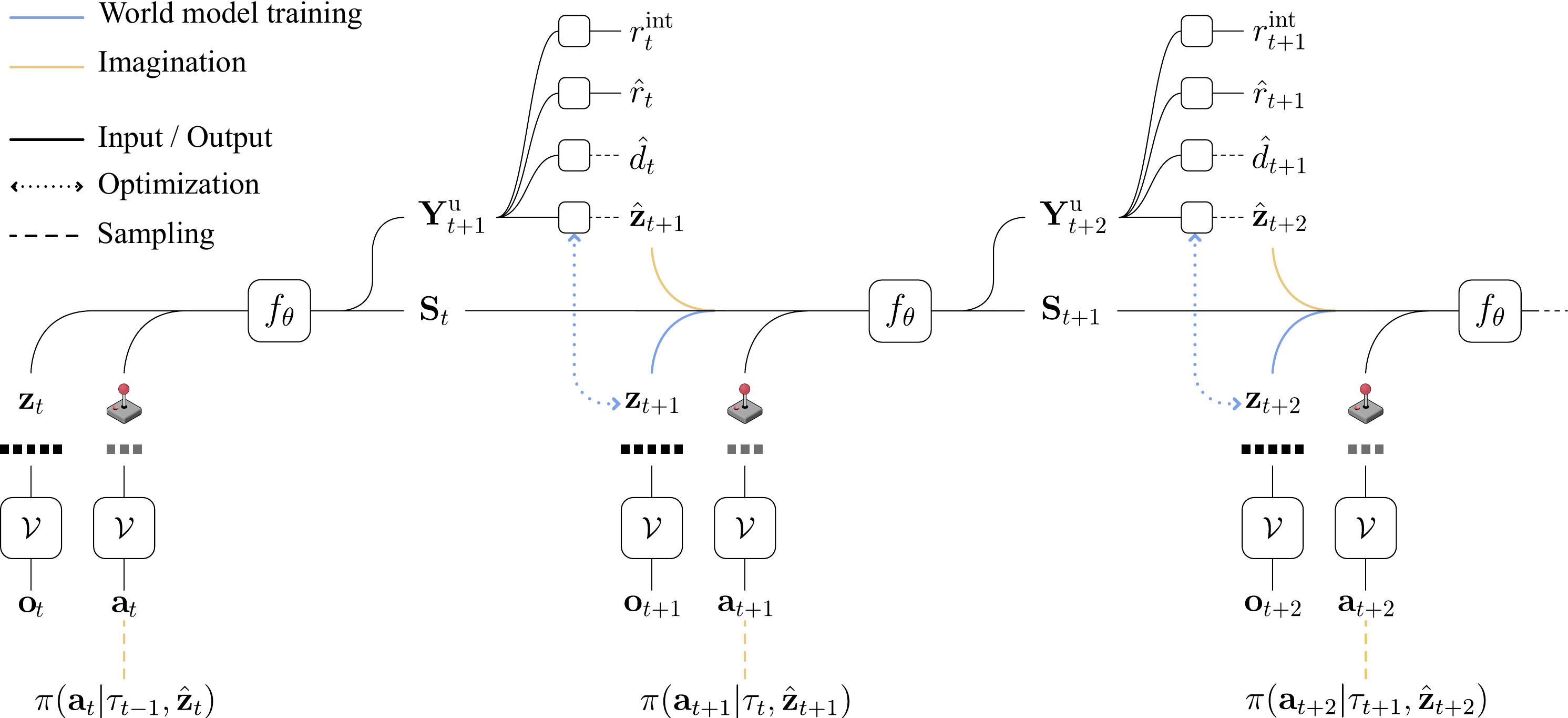}
    \caption{World model training and imagination. To maintain visual clarity, we omitted token embedding details, as well as optimization details of rewards and termination signals.}
    \label{fig:wm}
\end{figure*}

Similarly, rewards and termination signals are predicted by additional prediction heads as $\hat{\reward}_{t} = \rewardPredHead(\retnetOutputVec), \hat{\doneSgnl}_{t} \sim p_{\theta}( \hat{\doneSgnl}_{t} | \retnetOutputVec)$, slightly abusing notations, where $\retnetOutputVec$ is the last vector of $\retnetOutput^{\text{u}}_{t+1}$.
An illustration is provided in Figure \ref{fig:wm}.

\paragraph{Reducing Epistemic Uncertainty via Intrinsic Motivation}
The world model $\WM$ serves as the cornerstone of the entire system. 
Any controller operating within a world model framework can only perform as well as the underlying world model allows, making its quality a fundamental limiting factor. 
In turn, the model's performance depends heavily on the quality of its training data. 
Effective modeling requires data that sufficiently covers the environment's state-action space, yet achieving this coverage presents a particular challenge in online RL, where the controller must systematically and efficiently explore its environment. 
Success depends on intelligently guiding the controller toward unexplored or undersampled regions of the dynamics space. 
An effective approach to this challenge involves estimating the world model's epistemic uncertainty and directing the controller to gather data from regions where this uncertainty is highest \citep{Schmidhuber2010, pmlr-v97-shyam19a, sekar20aPlan2Explore}. 

Our approach estimates epistemic uncertainty using an ensemble of $\ensembleSize=4$ next observation prediction heads $\left\{ p_{\phi_i}(\hat{\tokens} | \stopGrad( \retnetOutput^{\text{u}} ) ) \right\}_{i=1}^{\ensembleSize}$ with parameters $\{ \phi_i \}_{i=1}^{\ensembleSize}$ \citep{sekar20aPlan2Explore, LakshminarayananNIPS2017uncertaintyEstDeepEns} where $\stopGrad(\cdot)$ is the stop gradient operator. 
To quantify disagreement between the ensemble's distributions, we employ the Jensen-Shannon divergence (JSD) \citep{pmlr-v97-shyam19a}. 
For probability distributions $P_1, \ldots, P_n$, the JSD is defined as:
\begin{equation*}
\JSD ( P_1, \ldots, P_n ) = \entropyFn(\frac{1}{n} \sum_{i=1}^{n} P_i ) - \frac{1}{n} \sum_{i=1}^{n} \entropyFn(P_i),
\end{equation*}
where $\entropyFn(\cdot)$ denotes the Shannon entropy. 
Since observations comprise multiple tokens, we average the per-token JSD values to obtain a single uncertainty measure $\disagreement_{t}$. 
Training data is divided equally among ensemble members, with each predictor processing a distinct subset of each batch.
Despite the ensemble approach, our implementation maintains computational efficiency, with negligible additional overhead in practice.



To guide $\Controller$ towards regions of high epistemic uncertainty, $\WM$ provides $\Controller$ with additional intrinsic rewards $\intReward_{t} = \disagreement_t$ during imagination.
Here, the reward provided by $\WM$ at each step $t$ is given by 
\begin{equation*}
    \bar{\reward}_t = \intRewardScale \intReward_t + \extRewardScale \extRewardPred_t ,
\end{equation*}
where $\intRewardScale, \extRewardScale \in \mathbb{R}$ are hyperparameters that control the scale of each reward type.
Optimizing the controller in imagination allows it to reach areas of high model uncertainty without additional real-environment interaction.

\paragraph{Prioritized Replay}
Recent work has demonstrated that prioritizing replay buffer sampling during world model training could lead to significant performance gains in intrinsically motivated agents \citep{kauvar2023curiousReplay}.
While their approach showed promise, it required extensive hyperparameter tuning in practice. 
We propose a simpler, more robust prioritization scheme for world model training.


Here, the replay buffer maintains a world model loss value for each stored example, with newly added examples assigned a high initial loss value $\replayBuffInitialValue$. During $\WM$'s training, we sample each batch using a mixture of uniform and prioritized sampling, controlled by a single parameter $\replayBufRatio \in [0, 1]$ that determines the fraction of prioritized samples. For the prioritized portion, we sample examples proportional to their softmax-transformed losses $p_i = \softmax(\mathcal{L})_i$.
The loss values are updated after each world model optimization step using the examples' current batch losses.


\paragraph{Training}
We use the cross-entropy loss for the optimization of all components of $\WM$.
Specifically, for each $t$, the loss of $p_{\theta}(\hat{\tokens}_{t} \vert \retnetOutput^{\text{u}}_{t})$ is obtained by averaging the cross-entropy losses of its individual tokens.
The same loss is used for each ensemble member $p_{\phi_i}(\hat{\tokens}_{t} \vert \stopGrad(\retnetOutput^{\text{u}}_{t}))$.
The optimization and design of the reward predictor is similar to that of the critic, as described in Section \ref{sec:classif-pred-cts-values}.
A formal description of the optimization objective can be found in Appendix \ref{sec:appendix-wm-optimization}.

\subsection{The Controller $\Controller$}
$\Controller$ is an extended version of the actor-critic of REM \citep{cohen2024improving} that supports additional observation and action spaces and implements regression-as-classification for return predictions.


\paragraph{Regression as Classification for Reward and Return Prediction}
\label{sec:classif-pred-cts-values}
Robustly handling unbounded reward signals has long been challenging as they can vary dramatically in both magnitude and frequency. 
\citet{hafner2023mastering} addressed this challenge by using a classification network that predicts the weights of exponentially spaced bins and employed a two-hot loss for the network's optimization. 
\citet{farebrother2024stop} studied the use of cross-entropy loss in place of the traditional mean squared error loss for value-based deep RL methods. 
In their work, the HL-Gauss method was shown to significantly outperform the two-hot loss method.
Building on these developments, we adopt the classification network with exponential bins from \citep{hafner2023mastering}, and apply the HL-Gauss method for its optimization.
Concretely, the critic's value estimates are predicted using a linear output layer 
parameterized by $\mathbf{W} \in \mathbb{R}^{m \times \CLSTMDim}$ with $m=128$ outputs corresponding to $m$ uniform bins defined by $m+1$ endpoints $\mathbf{b} = (b_0, \ldots, b_m)$.
The predicted value is given by
\begin{equation*}
    \hat{y} = \symexp\left(\softmax( \mathbf{W} \COutLatent )^{\Tr} \hat{\mathbf{b}} \right)
\end{equation*}
where $\COutLatent$ denotes the controller output latent representation (see Appendix~\ref{sec:controller-additional-details}), $\symexp(x) = \sign(x) (\exp(|x|) - 1)$ is the inverse of the symlog function, and $\hat{\mathbf{b}} = \left(\frac{b_1 + b_0}{2}, \ldots, \frac{b_{m} + b_{m-1}}{2}\right)$ are the bin centers.
Given the true target $y \in \mathbb{R}$, the HL-Gauss loss is given by
\begin{equation*}
    \mathcal{L}_{\text{HL-Gauss}}(\mathbf{W}, \COutLatent, y) = \tilde{\mathbf{y}}^{\Tr} \log  \softmax(\mathbf{W}\COutLatent)
\end{equation*}
where $\tilde{y}_i = \Phi(\frac{b_i - \symlog(y)}{\sigma}) - \Phi(\frac{b_{i-1} - \symlog(y)}{\sigma})$, $\Phi$ is the cumulative density function of the standard normal distribution and $\sigma$ is a standard deviation hyperparameter that controls the amount of label smoothing.

\paragraph{Training in Imagination}
$\Controller$ is trained entirely from simulated experience generated through interaction with $\WM$.
Specifically, $\WM$ and $\Controller$ are initialized with a short trajectory segment sampled uniformly from the replay buffer and interact for $\horizon$ steps.
An illustration of this process is given in Figure \ref{fig:wm} (orange path).
$\lambda$-returns are computed for each generated trajectory segment and are used as targets for critic learning.
For policy learning, a REINFORCE \citep{sutton1999REINFORCE} objective is used, with a $\valueFn$ baseline for variance reduction.
See Appendix \ref{sec:appendix-controller-optimization} for further details.

\section{Experiments}
\label{sec:experiments}



To evaluate sample efficiency, we used benchmarks that measure performance within a fixed, limited environment interaction budget.
These benchmarks were also selected to address key research questions: (1) whether the integrated components are individually effective and complementary across diverse environments; and (2) whether the proposed tokenization approach is effective in handling flexible combinations of observation and action modalities in sample-efficiency settings.

\subsection{Experimental Setup}

\paragraph{Benchmarks}
We evaluate \AlgName{} on three sample-efficiency benchmarks with diverse observation and action modalities:
Atari 100K \citep{Kaiser2020Model}, DMC Proprioception 500K \citep{tunyasuvunakool2020DMC, hafner2023mastering}, and Craftax-1M \citep{matthews2024craftax}, a partially observable 2D survival environment inspired by Minecraft.
These correspond to visual discrete control, continuous control from proprioceptive inputs, and mixed 2D symbolic and continuous observations, respectively.
Each benchmark imposes a strict interaction budget (100K, 500K, and 1M steps).
Additional environment details are provided in Appendix~\ref{sec:models-and-hyper-params}.

\paragraph{Baselines}
On Atari-100K, we compare \AlgName{} against DreamerV3 \citep{hafner2023mastering} and several methods restricted to image observations: TWM \citep{robine2023transformer}, STORM \citep{zhang2024storm}, DIAMOND \citep{alonso2024diffusion}, and REM \citep{cohen2024improving}.
On DMC, we compare exclusively with DreamerV3, currently the only planning-free world model method with published results on the 500K proprioception benchmark. 
On Craftax-1M, we compare against TWM \citep{dedieu2025TWM2}, a concurrent work that proposes a Transformer based world model with a focus on the Craftax benchmark, and the baselines reported in the Craftax paper: Random Network Distillation (RND) \citep{burda2018exploration}, PPO \citep{schulman2017proximal} with a recurrent neural network (PPO-RNN), and Exploration via Elliptical Episodic Bonuses (E3B) \citep{henaff2022explorationE3B}.
As Craftax is a recent benchmark, there are no other published results in existing world models literature.
Following the standard practice in the literature, we exclude planning-based methods \citep{hansen2024tdmpc, wang2024efficientzero}, as planning is an orthogonal component that operates on any given model, typically incurring significant computational overhead. 


\begin{figure*}[t]
    \centering
    
    \includegraphics[width=\linewidth]{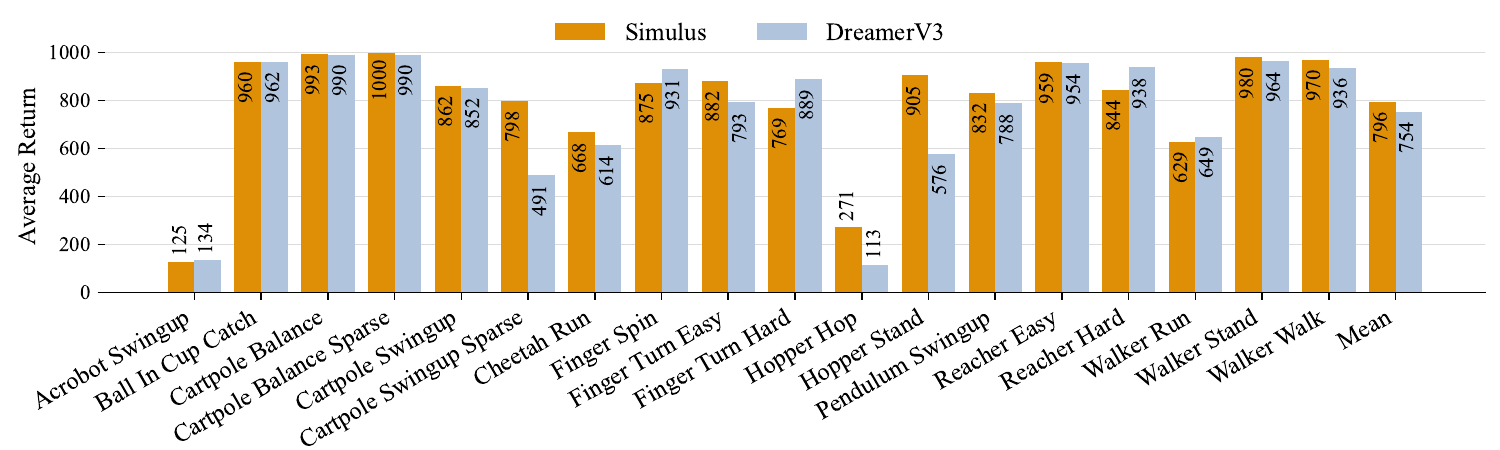}
    \includegraphics[width=\linewidth]{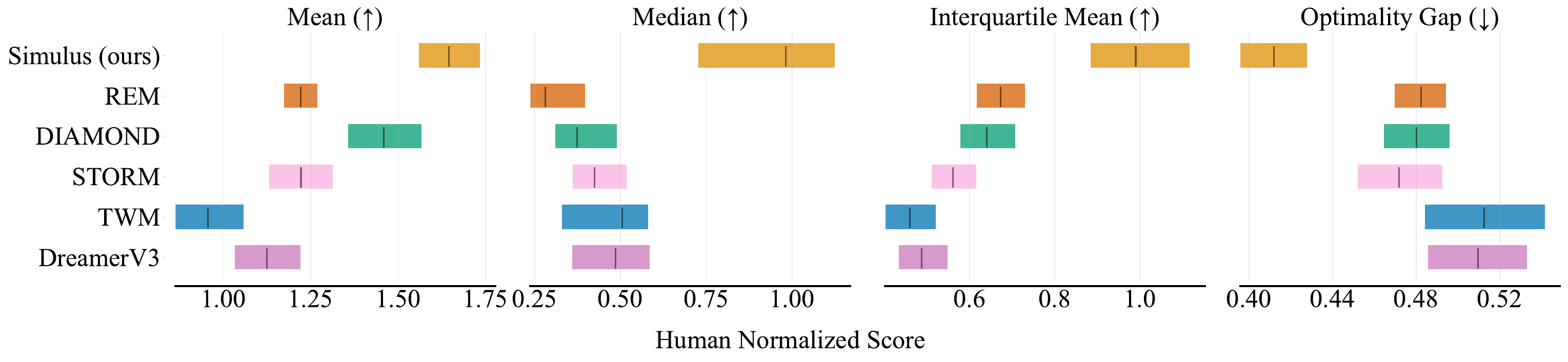}
    \caption{Results on the DeepMind Control Suite 500K Proprioception (top) and Atari 100K (bottom) benchmarks. }
    \label{fig:dmc-atari-aggregated-main-results}
\end{figure*}

\begin{wrapfigure}{r}{0.33\linewidth}
\vspace{0cm}
    \centering
    \includegraphics[width=\linewidth]{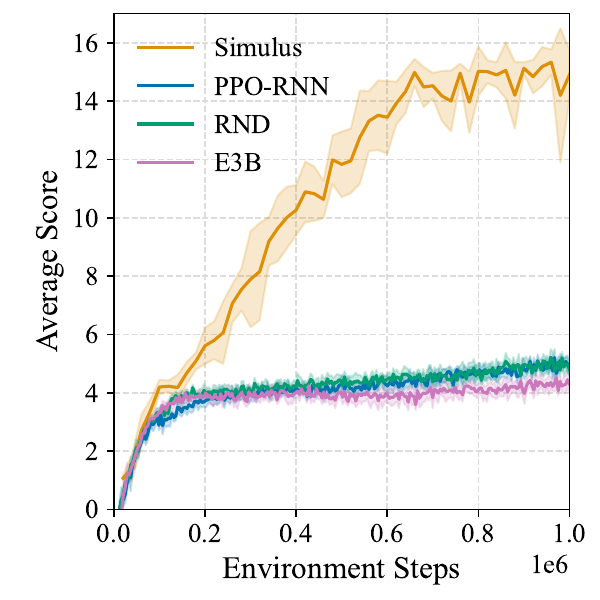}
    \caption{Craftax-1M training curves with mean and 95\% confidence intervals. }
    \label{fig:craftax-main-results}
\end{wrapfigure}

\paragraph{Metrics and Evaluation} 
For Atari, we report human-normalized scores (HNS), calculated as $\frac{\text{agent\_score} - \text{random\_score}}{\text{human\_score} - \text{random\_score}}$ \citep{mnih2015human}.
Following the protocol of \citet{Agarwal2021rliable} and using their toolkit, we report the mean, median, interquantile mean (IQM), and optimality gap metrics with 95\% stratified bootstrap confidence intervals.
For DMC and Craftax, we report the raw agent returns.
We use 5 random seeds per environment.
In each experiment, final performance is evaluated using 100 test episodes at the end of training and the mean score is reported.


\subsection{Results}


\AlgName{} achieves state-of-the-art performance across all three benchmarks (Figure~\ref{fig:first-page-results}).
On Atari 100k, it outperforms all baselines across key metrics (Figure~\ref{fig:dmc-atari-aggregated-main-results}).
Notably, \AlgName{} is the first planning-free world model to reach human-level IQM and median scores, achieving superhuman performance on \textbf{13} out of 26 games (Table~\ref{table:main-results}, Appendix~\ref{sec:appendix-additional-results}).
In Section \ref{sec:ablation-studies}, we show that these gains are attributed to the integration of the proposed components, demonstrating their individual and combined effectiveness.

\paragraph{Effectiveness in continuous control environments}
Figure~\ref{fig:dmc-atari-aggregated-main-results} provides compelling evidence that token-based architectures can perform well in continuous domains, even with compact vocabularies: \AlgName{} consistently matches DreamerV3 across most tasks and slightly outperforms it on average.

\paragraph{Effectiveness in environments with multi-modal observations}
We evaluate multi-modality performance in Craftax, as it combines a symbolic image-like grid map with a vector of continuous features, involving multiple tokenizers ($\Tokenizer$).
\AlgName{} maintains sample-efficiency in this multi-modal environment, outperforming both concurrent world model methods (Figure~\ref{fig:first-page-results}) and all model-free baselines (Figure~\ref{fig:craftax-main-results}), including exploration-focused algorithms.
With 444 tokens per observation arranged into sequences of 147 embeddings, even short trajectories in Craftax contain thousands of tokens, demonstrating \AlgName{}'s efficient handling of long sequences.
These findings indicate that the world model ($\WM$) and controller ($\Controller$) maintain strong performance under multi-modal inputs when processed by the proposed modular tokenizer.

\begin{figure}[t]
    \centering
    
\begin{subfigure}[t]{0.59\linewidth}
\vspace{0pt}
   \includegraphics[width=\linewidth]{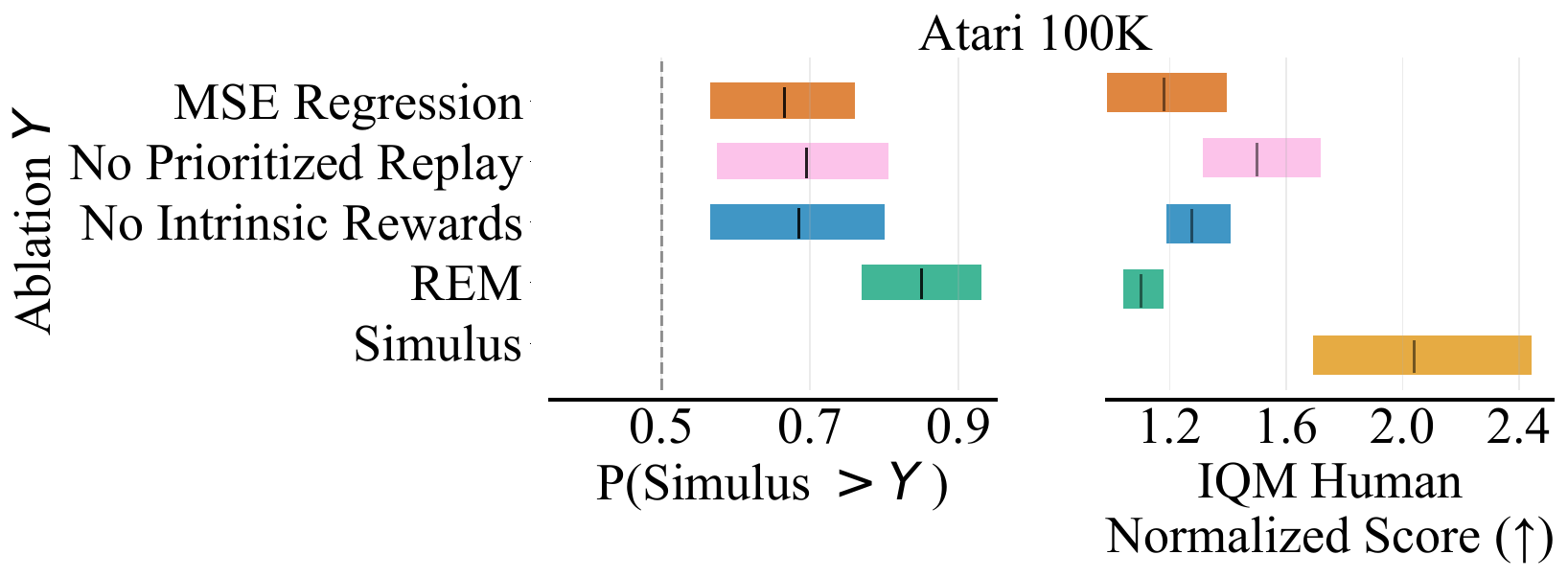}
   \label{fig:ablations-atari}
\end{subfigure}
\begin{subfigure}[t]{0.39\linewidth}
\vspace{0pt}
   \includegraphics[width=\linewidth]{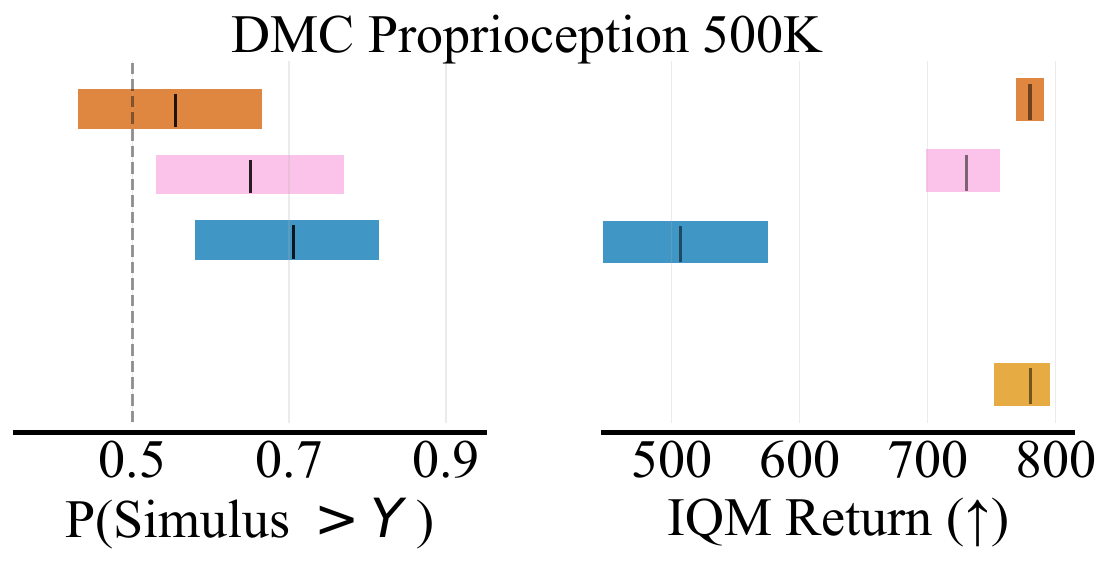}
   \label{fig:ablations-dmc}
\end{subfigure}
    
    \caption{Ablations results on the Atari-100K and DeepMind Control Proprioception 500K benchmarks. A subset of 8 games was used for each ablation.}
    \label{fig:ablations-results}
\end{figure}

\subsection{Ablation Studies}
\label{sec:ablation-studies}
We ablate the intrinsic rewards, prioritized replay, and regression-as-classification to demonstrate their individual contributions to \AlgName{}'s performance.
In each experiment, \AlgName{} is modified by disabling a single component.
Due to limited computational resources, we consider a subset of 8 tasks for each of the Atari and DMC benchmarks, and exclude Craftax from this analysis.
For Atari 100K, we report results on games exhibiting significant performance variation.
For DMC, we chose a subset that includes different embodiments.
We defer the specific environment names to Appendix \ref{sec:appendix-impl-details}.

The results are presented in Figure~\ref{fig:ablations-results}.
Although all components contributed to \AlgName{}'s final performance, intrinsic rewards were especially crucial for achieving competitive performance in both benchmarks.
Interestingly, the Atari 100K results indicate that combining all three components yields a significantly stronger algorithm.
These findings also suggest that both prioritized world model replay and regression-as-classification enhance the effectiveness of intrinsic rewards.

More broadly, the results in Figure~\ref{fig:ablations-results} demonstrate that encouraging the controller to explore regions of high epistemic uncertainty through intrinsic rewards significantly improves its performance in world model agents, even in reward-rich environments.
This observation is non-trivial in a sample-efficient setting, where the limited interaction budget makes model-driven exploration particularly costly, as it consumes resources that could otherwise be used for task-related exploration during data collection.
The latter type of exploration aims to collect new information about the true reward signal, which defines the task and its success metric.
On the other hand, model-driven exploration may guide the controller towards environment regions that are irrelevant to the task at hand.



\section{Related Work}
\paragraph{Offline Multi-Modal Methods}
Large-scale token sequence models for multi-modal agent trajectories have been proposed in \citep{lu2023unifiedio, Lu_2024_CVPR, reed2022a, schubert2023generalist}.
Gato \citep{reed2022a} and TDM \citep{schubert2023generalist} tokenize inputs via predefined transformations, while Unified IO \citep{lu2023unifiedio, Lu_2024_CVPR} leverages pretrained models. 
These methods do not learn control through RL but rely on expert data. 
They also use massive models with billions of parameters, large vocabularies, and significantly more data and compute than sample-efficient world models.
Consequently, it remains unclear whether their design choices would be effective in online, sample-efficient settings with non-stationary and limited data.

\paragraph{World Model Agents}


Model-based agents that learn policies solely from simulated data generated by a learned world model were introduced by \citet{ha2018worldmodels}, followed by the influential Dreamer family \citep{Hafner2020Dreamerv1, hafner2021dreamerv2, hafner2023mastering}. Dreamer jointly optimizes its representation and recurrent world models via a KL divergence between learned prior and posterior distributions, leading to interfering objectives (Appendix~\ref{sec:appendix-rssm-interference}) and a complex, monolithic architecture that complicates development and scaling \citep{wu2024ivideogpt}.
With the rise of Transformer architectures in language modeling \citep{NIPS2017_attn_is_all_you_need, brown2020LMsFewShotLearners}, Transformer-based Dreamer variants emerged \citep{zhang2024storm, robine2023transformer}, alongside token-based world models (TBWMs) that treat trajectories as discrete token sequences \citep{micheli2022transformers, cohen2024improving}. 
However, these methods are limited to visual environments with discrete actions (e.g., Atari 100K), leaving their performance in other modalities uncertain.
Recently, DIAMOND \citep{alonso2024diffusion}, a diffusion world model inspired by advances in generative modeling \citep{Rombach_2022_latentDiffusion}, was introduced. While it generates visually compelling outputs, it remains limited to visual domains.


\paragraph{Intrinsically Motivated World Model Agents}
Although intrinsic motivation (IM) has been extensively studied \citep{pathak2017curiosity, burda2018exploration, hazan2019provably, Badia2020Never, badia2020agent57}, its use in world models typically involves an exploration pretraining phase followed by limited task-specific fine-tuning \citep{sekar20aPlan2Explore, mendonca2021discovering, kauvar2023curiousReplay}. While combining IM with prioritized replay has shown promise \citep{kauvar2023curiousReplay}, it remains unexplored in standard sample-efficiency settings with external rewards.

\paragraph{Large-Scale Video World Models}
Building on recent advances in video generative modeling \citep{ho2022imagenVideo, blattmann2023stable, videoworldsimulators2024}, recent works have introduced large-scale video world models \citep{genie2, agarwal2025cosmos, oasis2024, valevski2024gameNGen}, trained offline on extensive pre-collected data to predict future frames. 
However, these models do not address control learning, particularly RL. 
While recent efforts aim to bridge this gap \citep{wu2024ivideogpt}, they remain confined to visual environments and lack comprehensive empirical evaluation.



\section{Limitations and Future Work}
\label{sec:limitations}
We briefly highlight several limitations of this work.
First, although the feature quantization approach for tokenizing continuous vectors showed promise, it leads to excessive sequence lengths.
We believe that more efficient solutions can be found for dealing with continuous inputs.
Second, due to the scarcity of rich multi-modal RL benchmarks, we could not extensively explore diverse modality combinations in our experiments.
Lastly, token-based world model agents remain slower to train than other baselines in sample-efficient RL.
Nonetheless, their modular design enables faster policy inference as the controller is independent of the world model.


\section{Conclusions}
\label{sec:conclusions}

Many promising improvements to world model agents remain unrealized, as combining them requires significant expertise and engineering effort. 
It is further unclear whether such improvements are even complementary, let alone synergistic.
Inspired by Rainbow, we introduce \AlgName{}: a modular token-based WM agent that combines multi-modality tokenization with promising but overlooked components: intrinsic motivation for epistemic uncertainty reduction, prioritized world model replay, and RaC with exponential binning for reward and return prediction.

Across visual (Atari 100K), continuous (DMC Proprioception 500K), and symbolic multi-modal (Craftax-1M) benchmarks, \AlgName{} achieves state-of-the-art sample efficiency for planning-free WMs, establishing TBWMs as versatile and widely applicable methods.
Ablation studies confirm that each component contributes individually and that their combination yields synergistic gains. 
Notably, world-model-induced exploration proves effective even under tight interaction budgets.


\bibliographystyle{plainnat}
\bibliography{references}

\newpage
\appendix

\section{Experimental Setup}
\label{sec:models-and-hyper-params}

\subsection{Benchmarks}
We evaluate \AlgName{} on three sample-efficiency benchmarks of different observation and action modalities:
Atari 100K \citep{Kaiser2020Model}, DeepMind Control Suite (DMC) Proprioception 500K \citep{tunyasuvunakool2020DMC}, and Craftax-1M \citep{matthews2024craftax}. 

Atari 100K has become the gold standard in the literature for evaluating sample-efficient deep RL agents.
The benchmark comprises a subset of 26 games.
Within each game, agents must learn from visual image signal under a tightly restricted budget of 100K interactions, corresponding to roughly two hours of human gameplay.

The DeepMind Control Suite (DMC) comprises continuous control tasks ranging from simple single-joint embodiments to complex humanoids. We follow the subset of proprioception tasks used to evaluate DreamerV3 \citep{hafner2023mastering}, featuring continuous observation and action vectors. Each task limits the agent's interaction budget to 500K steps.

Craftax is a 2D open-world survival game benchmark inspired by Minecraft, designed to evaluate RL agents' capabilities in planning, memory, and exploration. The partially-observable environment features procedurally generated worlds where agents must gather and craft resources while surviving against hostile creatures.
Observations consist of a 9×11 tile egocentric map, where each tile consists of 4 symbols, and 48 state features corresponding to state information such as inventory and health.
Here, we consider the sample-efficiency oriented Craftax-1M variant which only allows an interaction budget of one million steps.

\subsection{Hyperparameters}
We detail shared hyperparameters in \autoref{table:shared-hyperparams}, training hyperparameters in \autoref{table:training-hyperparams}, world model hyperparameters in \autoref{table:wm-hyperparams}, and controller hyperparameters in \autoref{table:controller-hyperparams}.
Environment hyperparameters are detailed in \autoref{table:atari-hyperparams} (Atari-100K) and \autoref{table:dmc-hyperparams} (DMC).

For the DMC benchmark, we use a lower embedding dimension (Table \ref{table:wm-hyperparams}) due to the significantly lower dimensionality of its observations compared to other benchmarks. As in prior work (e.g., DreamerV3), we adopt a smaller model for this setting. Note that the reduced number of Retention heads ensures a consistent head dimensionality (64).

Additionally, we used a limited decay range in DMC (Table \ref{table:wm-hyperparams}) as observations effectively represent full MDP states, eliminating the need for long-term memory. By constraining the decay range, we explicitly encode this inductive bias into the model.

In the Craftax benchmark, we reduce the number of layers (Table \ref{table:wm-hyperparams}) to lower computational cost. 
The interaction budget in Craftax is 1M steps, resulting in a particularly expensive training process. 
Here, we increase the decay range as an inductive bias to encourage long-term memory.

The weighting of intrinsic versus extrinsic rewards (Table~\ref{table:controller-hyperparams}) varies across benchmarks due to differences in reward structure and scale. For instance, DMC provides dense rewards with typical task scores reaching around 1000, whereas Craftax has extremely sparse rewards, with cumulative scores rarely exceeding 20—even over thousands of steps.

\paragraph{Tuning}
All remaining hyperparameters were tuned empirically or based on REM \citep{cohen2024improving}, with minimal impact on training cost and adjusted primarily for performance. 
Due to limited computational resources, we were unable to conduct extensive tuning, and we believe that further optimization could improve \AlgName{}'s performance.

\vfill

\begin{table}[h]
\caption{Shared hyperparameters.}
\label{table:shared-hyperparams}
\vskip 0.15in
\begin{center}
\begin{small}
\begin{tabular}{lccr}
\toprule
Description & Symbol & Value \\
\midrule
Eval sampling temperature &  & 0.5 \\
Optimizer &  & AdamW \\
Learning rate ($\Tokenizer$, $\WM$, $\Controller$) & & (1e-4, 2e-4, 2e-4) \\
AdamW $\beta_1$ &  & 0.9 \\
AdamW $\beta_2$ &  & 0.999 \\
Gradient clipping threshold ($\Tokenizer$, $\WM$, $\Controller$) & & (10, 3, 3) \\
Weight decay ($\Tokenizer$, $\WM$, $\Controller$) & & (0.01, 0.05, 0.01) \\

\midrule 

Prioritized replay fraction & $\alpha$ & 0.3 \\
Prioritized replay initial loss value & $\replayBuffInitialValue$ & 10 \\
$\WM$ ensemble size  & $\ensembleSize$ & 4 \\

\midrule

HL-Gauss num bins  &  & 128 \\
Label smoothing  &  $\sigma$  &  $\frac{3}{4} \text{bin\_width} = 0.1758 $ \\

\bottomrule
\end{tabular}
\end{small}
\end{center}
\vskip -0.1in
\end{table}

\begin{table}[h]
\caption{Training hyperparameters.}
\label{table:training-hyperparams}
\vskip 0.15in
\begin{center}
\begin{small}
\begin{tabular}{lccccr}
\toprule
Description & Symbol & Atari-100K & DMC & Craftax \\
\midrule
Horizon & $\horizon$ & 10 & 20 & 20 \\
Observation sequence length & $\tokensPerObs$ & 64 & 3-24 & 147 \\
Action sequence length & $\tokensPerObs_{\text{a}}$ & 1 & 1-6 & 1 \\
Tokenizer vocabulary size & $\tknzrVocabSize$ & 512 & 125 & (37,5,40,20,4,125) \\
\midrule
Epochs &  & 600 & 1000 & 10000 \\
Experience collection epochs &  & 500 & 1000 & 10000 \\
Environment steps per epoch &  & 200 & 500 & 100 \\
Batch size ($\Tokenizer$, $\WM$, $\Controller$) & & (128, 32, 128) & (-, 16, 128) & (-, 8, 128) \\ 
Training steps per epoch ($\Tokenizer$, $\WM$, $\Controller$) & & (200, 200, 80) & (-, 300, 100) & (-, 100, 50) \\
Training start after epoch ($\Tokenizer$, $\WM$, $\Controller$) & & (5, 25, 50) & (-, 15, 20) & (-, 250, 300) \\



\bottomrule
\end{tabular}
\end{small}
\end{center}
\vskip -0.1in
\end{table}

\begin{table}[h]
\caption{World model ($\WM$) hyperparameters.}
\label{table:wm-hyperparams}
\vskip 0.15in
\begin{center}
\begin{small}
\begin{tabular}{lccccr}
\toprule
Description & Symbol & Atari-100K & DMC & Craftax \\
\midrule
Number of layers &  & 10 & 10 & 5 \\
Number of heads & & 4 & 3 & 4 \\
Embedding dimension & $\retnetDmodel$ & 256 & 192 & 256 \\
Dropout & & 0.1& 0.1 & 0.1 \\
Retention decay range   & $[\retnetEta_{\text{min}}, \retnetEta_{\text{max}}]$ & [4, 16] & [2, 2] & [8, 40] \\



\bottomrule
\end{tabular}
\end{small}
\end{center}
\vskip -0.1in
\end{table}

\begin{table}[h]
\caption{Actor-critic ($\Controller$) hyperparameters.}
\label{table:controller-hyperparams}
\vskip 0.15in
\begin{center}
\begin{small}
\begin{tabular}{lccccr}
\toprule
Description & Symbol & Atari-100K & DMC & Craftax \\
\midrule
Environment reward weight & $\extRewardScale$ & 1 & 1 & 100 \\
Intrinsic reward weight & $\intRewardScale$ & 1 & 10 & 1 \\
Encoder MLP ($\CLatentFuser$) hidden layer sizes & & [512] & [384] & [512, 512] \\
Shared backbone & & True & False & True \\
Number of quantization values (continuous actions) & &  & 51 & \\
(2D) Categoricals embedding dimension & & & & 64 \\



\bottomrule
\end{tabular}
\end{small}
\end{center}
\vskip -0.1in
\end{table}

\pagebreak
\clearpage

\begin{table}[h]
\caption{Atari 100K hyperparameters.}
\label{table:atari-hyperparams}
\vskip 0.15in
\begin{center}
\begin{small}
\begin{tabular}{lccr}
\toprule
Description & Symbol & Value \\
\midrule

Frame resolution &  & $64 \times 64$ \\
Frame Skip &  & 4 \\
Max no-ops (train, test) &  & (30, 1) \\
Max episode steps (train, test) &  & (20K, 108K) \\
Terminate on live loss (train, test) &  & (No, Yes) \\

\bottomrule
\end{tabular}
\end{small}
\end{center}
\vskip -0.1in
\end{table}

\begin{table}[h]
\caption{DeepMind Control Suite Proprioception hyperparameters.}
\label{table:dmc-hyperparams}
\vskip 0.15in
\begin{center}
\begin{small}
\begin{tabular}{lccr}
\toprule
Description & Symbol & Value \\
\midrule
Action repeat &  & 2 \\

\bottomrule
\end{tabular}
\end{small}
\end{center}
\vskip -0.1in
\end{table}



\clearpage
\subsection{The Representation Module $\Tokenizer$}
\label{sec:repr-module-additional-details}

\subsubsection{Image Observations}
\label{sec:appendix-vqvae}
Image observations are tokenized using a vector-quantized variational auto-encoder (VQ-VAE) \citep{vanDenOord2017vqvae, esser2021tamingVQGAN}.
A VQ-VAE comprises a convolutional neural network (CNN) encoder, an embedding table $\embTable \in \mathbb{R}^{n \times \retnetDmodel}$, and a CNN decoder.
Here, the size of the embedding table $n$ determines the vocabulary size.

The encoder's output $\tknzrLatent \in \mathbb{R}^{W \times H \times \retnetDmodel}$ is a grid of $W \times H$ multi-channel vectors of dimension $\retnetDmodel$ that encode high-level learned features.
Each such vector is mapped to a discrete token by finding the closest embedding in $\embTable$:
\begin{equation*}
    \token = \arg\min_{i} \Vert \tknzrLatent - \embTable(i) \Vert ,
\end{equation*}
where $\embTable(i)$ is the $i$-th row of $\embTable$.
To reconstruct the original image, the decoder first maps $\tokens$ to their embeddings using $\embTable$.
During training, the straight-through estimator \citep{bengio2013estimating} is used for backpropagating the learning signal from the decoder to the encoder: $\hat{\tknzrLatent} = \tknzrLatent + \stopGrad(\embTable_{\token} - \tknzrLatent)$.
The architecture of the encoder and decoder models is presented in Table \ref{table:tokenizer-arch}.

The optimization objective is given by
\begin{multline*}
    \mathcal{L}(\encoder, \decoder, \embTable) = \\
    \Vert \obs - \decoder(z) \Vert_{2}^{2} + \Vert 
\stopGrad(\encoder(\obs)) - \embTable(z) \Vert_{2}^{2} + \Vert \stopGrad(\embTable(z) - \encoder(\obs) \Vert_{2}^{2} 
+ \mathcal{L}_{\text{perceptual}}(\obs, \decoder(z)) ,
\end{multline*}
where $\mathcal{L}_{\text{perceptual}}$ is a perceptual loss \citep{johnson2016perceptual, pmlr-v48-larsen16}, proposed in \citet{micheli2022transformers}.

Crucially, the learned embedding table $\embTable$ is used for embedding the (image) tokens across all stages of the algorithm.

\begin{table}[h]
\caption{The encoder and decoder architectures of the VQ-VAE model. ``Conv(a,b,c)" represents a convolutional layer with kernel size $a \times a$, stride of $b$ and padding $c$. A value of $c=\text{Asym.}$ represents an asymmetric padding where a padding of 1 is added only to the right and bottom ends of the image tensor. ``GN" represents a GroupNorm operator with $8$ groups, $\epsilon=1e-6$ and learnable per-channel affine parameters. SiLU is the Sigmoid Linear Unit activation \citep{hendrycks2017bridging, ramachandran2018searching}. ``Interpolate" uses PyTorch's interpolate method with scale factor of 2 and the ``nearest-exact" mode.}
\label{table:tokenizer-arch}
\vskip 0.15in
\begin{center}
\begin{small}
\begin{tabular}{lcr}
\toprule
Module & Output Shape \\
\midrule

Encoder \\
\midrule

Input & $3 \times 64 \times 64$ \\
Conv(3, 1, 1) & $64 \times 64 \times 64$ \\
EncoderBlock1 & $128 \times 32 \times 32$ \\
EncoderBlock2 & $256 \times 16 \times 16$ \\
EncoderBlock3 & $512 \times 8 \times 8$ \\
GN & $512 \times 8 \times 8$ \\
SiLU & $512 \times 8 \times 8$ \\
Conv(3, 1, 1) & $256 \times 8 \times 8$ \\

\midrule

EncoderBlock  \\
\midrule
Input & $c \times h \times w$ \\
GN & $c \times h \times w$ \\
SiLU & $c \times h \times w$ \\
Conv(3, 2, \text{Asym.}) & $2c \times \frac{h}{2} \times \frac{w}{2}$ \\

\midrule
Decoder \\
\midrule
Input & $256 \times 8 \times 8$ \\
BatchNorm & $256 \times 8 \times 8$ \\
Conv(3, 1, 1) & $256 \times 8 \times 8$ \\
DecoderBlock1 & $128 \times 16 \times 16$ \\
DecoderBlock2 & $64 \times 32 \times 32$ \\
DecoderBlock3 & $64 \times 64 \times 64$ \\
GN & $64 \times 64 \times 64$ \\
SiLU & $64 \times 64 \times 64$ \\
Conv(3, 1, 1) & $3 \times 64 \times 64$ \\

\midrule
DecoderBlock \\
\midrule
Input & $c \times h \times w$ \\
GN & $c \times h \times w$ \\
SiLU & $c \times h \times w$ \\
Interpolate & $c \times 2h \times 2w$ \\
Conv(3, 1, 1) & $\frac{c}{2} \times 2h \times 2w$ \\

\bottomrule
\end{tabular}
\end{small}
\end{center}
\vskip -0.1in
\end{table}


\subsubsection{Continuous Vectors}
\label{sec:appendix-cts-vectors-tokenization}
For the tokenization of continuous vectors, unbounded inputs are first transformed using the symlog function \citep{hafner2023mastering}, defined as:
\begin{equation*}
    \text{symlog}(x) = \text{sign}(x)\ln(1 + |x|) .
\end{equation*}
This transformation compresses the magnitude of large absolute values. Then, each feature is quantized to produce a discrete token, following \citep{reed2022a}.

To improve learning efficiency, we reduce the vocabulary size from 1024 to 125 and modify the quantization levels for optimal coverage. Specifically, the quantization of each feature uses $125$ values in the range $[-6, 6]$, where $63$ values are uniformly distributed in $[-\ln(1+\pi), \ln(1+\pi)]$ and the rest are uniformly distributed in the remaining intervals.

Lastly, while no special tokenization is required for categorical inputs, 2D categoricals are flattened along the spatial dimensions to form a sequence of categorical vectors. The embedding of each token vector is obtained by averaging the embeddings of its entries.

\clearpage
\subsection{The World Model $\WM$}
\label{sec:wm-additional-details}

\paragraph{Embedding Details}
Each token in $\tokens^{(i)}$ of each modality is mapped to a $\retnetDmodel$-dimensional embedding vector $\tknEmbs^{(i)}$ using the embedding (look-up) table $\embTable^{(i)}$ of modality $\modalitySet_i$.
The embedding vector that corresponds to token $\token$ is simply the $\token$-th row in the embedding table.
Formally, $\tknEmb^{(i)}_{t, j} = \embTable^{(i)}(l),\  l=\token^{(i)}_{t, j}$ where $\embTable(l)$ refers to the $l$-th row in $\embTable$.
In the special case of 2D categorical inputs, $\tknEmb^{(i)}_{t, j} = \frac{1}{C} \sum_{n=1}^{C} \embTable^{(i)}_{n}(l_n), \ l_n = \token^{(i)}_{t, j, n}$ where $C$ is the number of channels and $i$ is the index of the 2D categorical modality in $\modalitySet$.

To concatenate the embeddings, we use the following order among the modalities: images, continuous vectors, categorical variables, and 2D categoricals.

\paragraph{Prediction Heads}
Each prediction head in $\WM$ is a multi-layer perceptron (MLP) with a single hidden layer of dimension 2$\retnetDmodel$ where $\retnetDmodel$ is the embedding dimension.

\paragraph{Epistemic Uncertainty Estimation}
Working with discrete distributions enables efficient entropy computation and ensures that the ensemble disagreement term $\disagreement_t$ is bounded by $\frac{1}{|\tokens_{t}|} \sum_{\token \in \tokens_{t}} \log(\vocabSize(z))$.


\subsubsection{Optimization}
\label{sec:appendix-wm-optimization}
For each training example in the form of a trajectory segment in token representation $\tau = \tokens_{1}, \actionTokens_{1}, \ldots, \tokens_{H}, \actionTokens_{H}$, the optimization objective is given by
\begin{multline*}
    \mathcal{L}_{\WM}(\theta, \phi, \tau) = 
    \sum_{t=1}^{H}  \mathcal{L}_{\text{obs}}(\theta, \tokens_{t}, p_{\theta}(\hat{\tokens}_{t} \vert \retnetOutput^{\text{u}}_{t})) 
    + \mathcal{L}_{\text{reward}}(\theta, \reward_{t}, \hat{\reward}_{t}) 
    -\log(p_{\theta}({\doneSgnl}_{t} \vert \retnetOutput^{\text{u}}_{t})) \\
    + \sum_{i=1}^{\ensembleSize} \mathcal{L}_{\text{obs}}(\phi_i, \tokens_{t}, p_{\phi_i}(\hat{\tokens}_{t} \vert \stopGrad( \retnetOutput^{\text{u}}_{t})) ) ,
\end{multline*}
where
\begin{equation*}
    \mathcal{L}_{\text{obs}}(\theta, \tokens_{t}, p_{\theta}(\hat{\tokens}_{t} \vert \retnetOutput^{\text{u}}_{t})) = -\frac{1}{\tokensPerObs} \sum_{i=1}^{\tokensPerObs} \log p_{\theta}(z_i | \retnetOutputVec_i)
\end{equation*}
is the average of the cross-entropy losses of the individual tokens, and $\mathcal{L}_{\text{reward}}(\theta, \reward_{t}, \hat{\reward}_{t})$ is the $\mathcal{L}_{\text{HL-Gauss}}$ loss with the respective parameters of the reward head.
Here, $\retnetOutputVec_i$ is the vector of $\retnetOutput^{\text{u}}_{t}$ that corresponds to $\token_i$, the $i$-th token of $\tokens_{t}$.

\subsubsection{Retentive Networks}
Retentive Networks (RetNet) \citep{sun2023retentive} are sequence model architectures with a Transformer-like structure \citep{NIPS2017_attn_is_all_you_need}.
However, RetNet replaces the self-attention mechanism with a linear-attention \citep{pmlr-v119-katharopoulos20a} based Retention mechanism.
At a high level, given an input sequence $\tknEmbs \in \mathbb{R}^{|\tknEmbs| \times \retnetDmodel}$ of $\retnetDmodel$-dimensional vectors, the Retention operator outputs
\begin{equation*}
    \Retention(\tknEmbs) = (\retnetQ \retnetK^{\Tr} \odot \retnetD) \retnetV ,
\end{equation*}
where $\retnetQ, \retnetK, \retnetV$ are the queries, keys, and values, respectively, and $\retnetD$ is a causal mask and decay matrix.
Notably, the softmax operation is discarded in Retention and other linear attention methods.
As a linear attention method, the computation can also be carried in a recurrent form:
\begin{equation*}
\begin{split}
    \Retention(\tknEmb_{t}, \retnetState_{t-1}) & = \retnetState_{t} \retnetQVec_{t} , \\
    \retnetState_{t} & = \retnetEta \retnetState_{t-1} + \retnetVVec_{t} \retnetKVec_{t}^{\Tr} \in \mathbb{R}^{\retnetDmodel \times \retnetDmodel} ,
\end{split}
\end{equation*}
where $\retnetEta$ is a decay factor, $\retnetState_{t}$ is a recurrent state, and $\retnetState_{0} = 0$.
In addition, a hybrid form of recurrent and parallel forward computation known as the chunkwise mode allows to balance the quadratic cost of the parallel form and the sequential cost of the recurrent form by processing the input as a sequence of chunks.
We refer the reader to \citet{sun2023retentive} for the full details about this architecture.

In our implementation, since inputs are complete observation-action block sequences $\tknEmbs_{1},\ldots, \tknEmbs_{t}$, we configure the decay factors of the multi-scale retention operator in block units:
\begin{gather*}
    \retnetEta = 1 - 2^{ - \linspace(\log_2(\tokensPerObs \retnetEta_{\text{min}}), \log_2(\tokensPerObs  \retnetEta_{\text{max}}  ), N_h )    } ,
\end{gather*}
where $\linspace(a, b, c )$ is a sequence of $c$ values evenly distributed between $a$ and $b$, $N_h$ is the number of retention heads, and $\retnetEta_{\text{min}}, \retnetEta_{\text{max}}$ are hyperparameters that control the memory decay in observation-action block units.

\subsubsection{Parallel Observation Prediction (POP)}
POP \citep{cohen2024improving} is a mechanism for parallel generation of non-causal subsequences such as observations in token representation.
It's purpose is to improve generation efficiency by alleviating the sequential bottleneck caused by generating observations a single token at a time (as done in language models).
However, to achieve this goal, POP also includes a mechanism for maintaining training efficiency.
Specifically, POP extends the chunkwise forward mode of RetNet to maintain efficient training of the sequence model.

To generate multiple tokens into the future at once, POP introduces a set of prediction tokens $\predTokens = \predToken_{1}, \ldots, \predToken_{\tokensPerObs}$ and embeddings $\predEmb \in \mathbb{R}^{\tokensPerObs \times \retnetDmodel}$ where $\tokensPerObs$ is the number of tokens in an observation.
Each token in $\predTokens$ corresponds to an observation token in $\tokens$.
These tokens, and their respective learned embeddings, serve as a learned prior.

Let $\tknEmbs_{1}, \ldots, \tknEmbs_{T}$ be a sequence of $T$ observation-action (embeddings) blocks.
Given $\retnetState_{t-1}$ summarizing all key-value outer products of elements of $\tknEmbs_{\leq t-1}$, the outputs $\retnetOutput^{\text{u}}$ from which the next observation tokens are predicted are given by:
\begin{equation*}
    (\cdot, \retnetOutput^{\text{u}}_{t}) = \seqModel(\retnetState_{t-1}, \predEmb) .
\end{equation*}
Importantly, the recurrent state is never updated based on the prediction tokens $\predTokens$ (or their embeddings).
The next observation tokens $\hat{\tokens}_{t}$ are sampled from $p_{\theta}(\hat{\tokens}_{t} | \retnetOutput^{\text{u}}_{t})$.
Then, the next action is generated by the controller, and the next observation-action block $\tknEmbs_{t}$ can be processed to predict the next observation $\hat{\tokens}_{t+1}$.

To maintain efficient training, a two step computation is carried at each RetNet layer.
First, all recurrent states $\retnetState_{t}$ for all $1\leq t \leq T$ are calculated in parallel. Although there is an auto-regressive relationship between time steps, the linear structure of $\retnetState$ allows to calculate the compute-intensive part of each state in parallel and incorporate past information efficiently afterwards.
In the second step, all outputs $\retnetOutput^{\text{u}}_{t}$ for all $1 \leq t \leq T$ are computed in parallel, using the appropriate states $\retnetState_{t-1}$ and $\tknEmbs^{\text{u}}$ in batch computation.
Note that this computation involves delicate positional information handling.
We refer the reader to \citet{cohen2024improving} for full details of this computation.

    


\newpage
\subsection{The Controller $\Controller$}
\label{sec:controller-additional-details}

\paragraph{Architecture} 
At the core of $\Controller$'s architecture, parameterized by $\CParams$, is an LSTM \citep{hochreiter1997lstm} sequence model.
At each step $t$, upon observing $\tokens_{t}$, a set of modality-specific encoders map each modality tokens $\tokens^{(i)}_{t}$ to a latent vector $\CLatent^{(i)}$, where we abuse our notation $\tknEmb$ as the context of the discussion is limited to $\Controller$.
The latents are then fused by a fully-connected network to obtain a single vector $\CLatent = \CLatentFuser(\CLatent^{(1)}, \ldots, \CLatent^{(|\modalitySet|)})$.
$\CLatent_{t} \in \mathbb{R}^{\CLSTMDim}$ is processed by $\Controller$'s sequence model to produce $\COutLatent_{t}, \CHiddenState_{t} = \LSTM(\CLatent_{t}, \COutLatent_{t-1}, \CHiddenState_{t-1}; \CParams)$ where $\COutLatent_{t}, \CHiddenState_{t}$ are the LSTM's hidden and cell states, respectively.
Lastly, two linear output layers produce the logits from which the actor and critic outputs $\policy( \action_{t} | \COutLatent_{t}), \valueFn(\COutLatent_{t})$ are derived.
For continuous action spaces, the actor uses a categorical distribution over a uniformly spaced discrete subset of $[-1, 1]$.

\paragraph{Critic}
The value prediction uses 128 bins in the range $\mathbf{b} = (-15, \ldots, 15)$.

\paragraph{Continuous Action Spaces}
The policy network outputs $m=51$ logits corresponding to $m$ quantization values uniformly distributed in $[-1, 1]$ for each individual action in the action vector.


\subsubsection{Input Encoding}
\label{sec:appendix-C-input-encoding}
The controller $\Controller$ operates in the latent token space.
Token trajectories $\tau = \tokens_{1}, \actionTokens_{1}, \ldots, \tokens_{H}, \actionTokens_{H}$ are processed sequentially by the LSTM model.
At each time step $t$, the network gets $\tokens_{t}$ as input, outputs $\policy(\action_{t} \vert \tau_{\leq t-1}, \tokens_{t})$ and $\valueFn(\action_{t} \vert \tau_{\leq t-1}, \tokens_{t})$, samples an action $\action_{t}$ and then process the sampled action as another sequence element.

The processing of actions involve embedding the action into a latent vector which is then provided as input to the LSTM.
Embedding of continuous action tokens is performed by first reconstructing the continuous action vector and then computing the embedding using a linear projection.
Discrete tokens are embedded using a dedicated embedding table.


To embed observation tokens $\tokens$, the tokens of each modality are processed by a modality-specific encoder.
The outputs of the encoders are concatenated and further processed by a MLP $\CLatentFuser$ that combines the information into a single vector latent representation.

The image encoder is a convolutional neural network (CNN). Its architecture is given in \autoref{table:actor-critic-obs-encoder-arch}.

Categorical variables are embedded using a learned embedding table.
For 2D categoricals, shared per-channel embedding tables map tokens to embedding vectors, which are averaged to obtain a single embedding for each multi-channel token vector. 
For both types of categorical inputs we use 64 dimensional embeddings.
The embeddings are concatenated and processed by $\CLatentFuser$.

\begin{table}[h]
\caption{The image observation encoder architecture of the actor-critic controller $\Controller$.}
\label{table:actor-critic-obs-encoder-arch}
\vskip 0.15in
\begin{center}
\begin{small}
\begin{tabular}{lcr}
\toprule
Module & Output Shape \\
\midrule

Input & $256 \times 8 \times 8$ \\
Conv(3, 1, 1) & $128 \times 8 \times 8$ \\
SiLU & $128 \times 8 \times 8$ \\
Conv(3, 1, 1) & $64 \times 8 \times 8$ \\
SiLU & $64 \times 8 \times 8$ \\
Flatten & 4096 \\
Linear & 512 \\
SiLU & 512 \\

\bottomrule
\end{tabular}
\end{small}
\end{center}
\vskip -0.1in
\end{table}

\subsubsection{Optimization}
\label{sec:appendix-controller-optimization}
$\lambda$-returns are computed for each generated trajectory segment $\hat{\tau} = (\tokens_{1}, \action_{1}, \bar{\reward}_{1}, \doneSgnl_{1}, \hat{\tokens}_{2}, \action_{2}, \bar{\reward}_{2}, \doneSgnl_{2}, \ldots, \hat{\tokens}_{H}, \action_{H}, \bar{\reward}_{H}, \doneSgnl_{H})$:
\begin{equation*}
    G_{t} = \begin{cases}
        \bar{\reward}_{t} + \discountF (1-\doneSgnl_t)( (1-\lambda) \valueFn_{t+1} + \lambda G_{t+1} ) & t<H \\
        \valueFn_{H} & t=H
    \end{cases} ,
\end{equation*}
where $\valueFn_{t} = \valueFn(\hat{\tau}_{\leq t})$.
These $\lambda$-returns are used as targets for critic learning.
For policy learning, a REINFORCE \citep{sutton1999REINFORCE} objective is used, with a normalized $\valueFn$ baseline for variance reduction:
\begin{equation*}
    \mathcal{L}_{\policy}(\CParams) = \E_{\policy}\left[ \sum_{t=1}^{H} \stopGrad\left( \frac{G_{t} - \valueFn_{t}}{\max(1, c)} \right) \log \policy(\action_{t} \vert \hat{\tau}_{\leq t-1}, \hat{\tokens}_{t})  + \CentropyWeight \mathcal{H}(\policy(\action_{t} \vert \hat{\tau}_{\leq t-1}, \hat{\tokens}_{t})) \right] ,
\end{equation*}
where $c$ is an estimate of the effective return scale similar to DreamerV3 \citep{hafner2023mastering} and $\CentropyWeight$ is a hyperparameter that controls the entropy regularization weight.
$c$ is calculated as the difference between the running average estimators of the $97.5$ and $2.5$ return percentiles, based on a window of return estimates obtained in the last $500$ batches (imagination).

\newpage
\section{Additional Results}
\label{sec:appendix-additional-results}
\begin{figure}[h]
    \centering
    \includegraphics[width=0.7\linewidth]{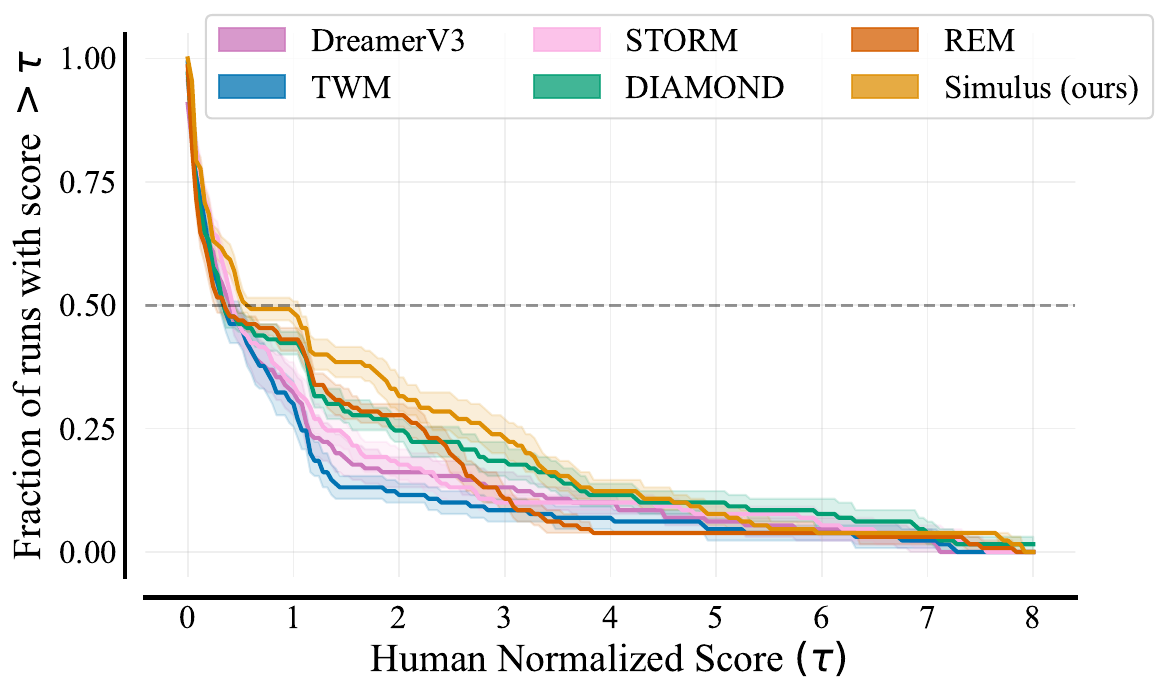}
    \caption{Performance profile. For each human-normalized score value on the x-axis, the curve of each algorithm represents the fraction of runs achieving a score greater than that value. Shaded regions denote pointwise 95\% confidence intervals, computed using stratified bootstrap sampling \citep{Agarwal2021rliable}. }
    \label{fig:performance-profile-atari}
\end{figure}

The average per-game scores for Atari-100K are presented in Table \ref{table:main-results}.
The performance profile plot for Atari 100K is presented in Figure \ref{fig:performance-profile-atari}.

\begin{table*}[h]
\caption{Mean returns on the 26 games of the Atari 100k benchmark followed by averaged human-normalized performance metrics. Each game score is computed as the average of 5 runs with different seeds. Bold face mark the best score.}
\label{table:main-results}
\begin{center}
\begin{scriptsize}
\begin{tabular}{lcc ccccc cr}
\toprule

Game                 &  Random    &  Human     &  DreamerV3          &  TWM                &  STORM              &  DIAMOND            &  REM               &  \textsc{\AlgName} (ours)  \\
\midrule
Alien                &  227.8     &  7127.7    &  959.4              &  674.6              &  \textbf{983.6}     &  744.1              &  607.2             &  687.2                 \\
Amidar               &  5.8       &  1719.5    &  139.1              &  121.8              &  204.8              &  \textbf{225.8}     &  95.3              &  102.4                 \\
Assault              &  222.4     &  742.0     &  705.6              &  682.6              &  801.0              &  1526.4             &  1764.2            &  \textbf{1822.8}       \\
Asterix              &  210.0     &  8503.3    &  932.5              &  1116.6             &  1028.0             &  \textbf{3698.5}    &  1637.5            &  1369.1                \\
BankHeist            &  14.2      &  753.1     &  \textbf{648.7}     &  466.7              &  641.2              &  19.7               &  19.2              &  347.1                 \\
BattleZone           &  2360.0    &  37187.5   &  12250.0            &  5068.0             &  \textbf{13540.0}   &  4702.0             &  11826.0           &  13262.0               \\
Boxing               &  0.1       &  12.1      &  78.0               &  77.5               &  79.7               &  86.9               &  87.5              &  \textbf{93.5}         \\
Breakout             &  1.7       &  30.5      &  31.1               &  20.0               &  15.9               &  132.5              &  90.7              &  \textbf{148.9}        \\
ChopperCommand       &  811.0     &  7387.8    &  410.0              &  1697.4             &  1888.0             &  1369.8             &  2561.2            &  \textbf{3611.6}       \\
CrazyClimber         &  10780.5   &  35829.4   &  97190.0            &  71820.4            &  66776.0            &  \textbf{99167.8}   &  76547.6           &  93433.2               \\
DemonAttack          &  152.1     &  1971.0    &  303.3              &  350.2              &  164.6              &  288.1              &  \textbf{5738.6}   &  4787.6                \\
Freeway              &  0.0       &  29.6      &  0.0                &  24.3               &  0.0                &  \textbf{33.3}      &  32.3              &  31.9                  \\
Frostbite            &  65.2      &  4334.7    &  909.4              &  \textbf{1475.6}    &  1316.0             &  274.1              &  240.5             &  258.4                 \\
Gopher               &  257.6     &  2412.5    &  3730.0             &  1674.8             &  \textbf{8239.6}    &  5897.9             &  5452.4            &  4363.2                \\
Hero                 &  1027.0    &  30826.4   &  \textbf{11160.5}   &  7254.0             &  11044.3            &  5621.8             &  6484.8            &  7466.8                \\
Jamesbond            &  29.0      &  302.8     &  444.6              &  362.4              &  509.0              &  427.4              &  391.2             &  \textbf{678.0}        \\
Kangaroo             &  52.0      &  3035.0    &  4098.3             &  1240.0             &  4208.0             &  5382.2             &  467.6             &  \textbf{6656.0}       \\
Krull                &  1598.0    &  2665.5    &  7781.5             &  6349.2             &  8412.6             &  \textbf{8610.1}    &  4017.7            &  6677.3                \\
KungFuMaster         &  258.5     &  22736.3   &  21420.0            &  24554.6            &  26182.0            &  18713.6            &  25172.2           &  \textbf{31705.4}      \\
MsPacman             &  307.3     &  6951.6    &  1326.9             &  1588.4             &  \textbf{2673.5}    &  1958.2             &  962.5             &  1282.7                \\
Pong                 &  -20.7     &  14.6      &  18.4               &  18.8               &  11.3               &  \textbf{20.4}      &  18.0              &  19.9                  \\
PrivateEye           &  24.9      &  69571.3   &  881.6              &  86.6               &  \textbf{7781.0}    &  114.3              &  99.6              &  100.0                 \\
Qbert                &  163.9     &  13455.0   &  3405.1             &  3330.8             &  \textbf{4522.5}    &  4499.3             &  743.0             &  2425.6                \\
RoadRunner           &  11.5      &  7845.0    &  15565.0            &  9109.0             &  17564.0            &  20673.2            &  14060.2           &  \textbf{24471.8}      \\
Seaquest             &  68.4      &  42054.7   &  618.0              &  774.4              &  525.2              &  551.2              &  1036.7            &  \textbf{1800.4}       \\
UpNDown              &  533.4     &  11693.2   &  7567.1             &  \textbf{15981.7}   &  7985.0             &  3856.3             &  3757.6            &  10416.5               \\
\midrule
\#Superhuman (↑)     &  0         &  N/A       &  9                  &  8                  &  9                  &  11                 &  12                &  \textbf{13}           \\
Mean (↑)             &  0.000     &  1.000     &  1.124              &  0.956              &  1.222              &  1.459              &  1.222             &  \textbf{1.645}        \\
Median (↑)           &  0.000     &  1.000     &  0.485              &  0.505              &  0.425              &  0.373              &  0.280             &  \textbf{0.982}        \\
IQM (↑)              &  0.000     &  1.000     &  0.487              &  0.459              &  0.561              &  0.641              &  0.673             &  \textbf{0.990}        \\
Optimality Gap (↓)   &  1.000     &  0.000     &  0.510              &  0.513              &  0.472              &  0.480              &  0.482             &  \textbf{0.412}        \\
\bottomrule
\end{tabular}
\end{scriptsize}
\end{center}
\end{table*}

\vfill

\pagebreak

\subsection{Interfering Objectives in RSSM Optimization}
\label{sec:appendix-rssm-interference}
Here, we study the interplay between the objectives of a Dreamer-like world model, PWM \citep{wang2024parallelizing}, which uses a slightly modified version of the recurrent state space model (RSSM) of Dreamer.
Concretely, we aim to understand whether the representation and sequence modeling objectives interfere by decoupling the optimization of the encoder-decoder models from that of the world model.
We opted for this implementation due to its simplicity, fast runtime, and accessibility, as it is written in PyTorch \citep{Paszke2019Pytorch}.

Formally, the model consists of the following components:
\begin{align*}
    \text{Encoder:}\quad & z_t \sim q_{\theta}(z_t | o_t), \\
    \text{Decoder:}\quad & \hat{o}_t \sim p_{\theta}(\hat{o}_t | z_t), \\
    \text{Sequence model:}\quad & h_t, x_t = f_{\theta}(x_{t-1}, z_{t-1}, a_{t-1}), \\ 
    \text{Dynamics predictor:}\quad & \hat{z}_t \sim p_{\theta}(\hat{z}_t | h_t). 
\end{align*}
We omit the reward and termination predictors and objectives for brevity.
Note that in contrast to the RSSM in Dreamer, the encoder and decoder models do not depend on the recurrent state $h_t, x_t$.
The optimization objective of PWM is given by 
\begin{equation*}
    \mathcal{L}(\theta) = \E_{q_{\theta}} [ \sum_{t=1}^{T} \beta_{\text{pred}} \mathcal{L}_{\text{pred}}(\theta) + \beta_{\text{dyn}}\mathcal{L}_{\text{dyn}}(\theta) + \beta_{\text{rep}}\mathcal{L}_{\text{rep}}(\theta) ] ,
\end{equation*}
where $\beta_{\text{pred}}, \beta_{\text{dyn}}$, and $\beta_{\text{rep}}$ are coefficients and
\begin{align*}
    \mathcal{L}_{\text{pred}}(\theta) = & \Vert \hat{o}_t - o_t \Vert_{2}^{2} , \\
    \mathcal{L}_{\text{dyn}}(\theta) = & \max(1, \KLD[\stopGrad(q_{\theta}(z_t | o_t)) \Vert p_{\theta}(\hat{z}_t | h_t) ]) , \\
    \mathcal{L}_{\text{rep}}(\theta) = & \max(1, \KLD[q_{\theta}(z_t | o_t) \Vert \stopGrad((p_{\theta}(\hat{z}_t | h_t)) ])  .
\end{align*}

To decouple the optimization, we modify the sequence model by introducing a stop-gradient operator on the encoder's output during world model training:
\begin{equation*}
    h_t, x_t = f_{\theta}(x_{t-1}, \stopGrad(z_{t-1}), a_{t-1}) .
\end{equation*}
Moreover, this modification allows to train the encoder-decoder models using large batches of single frames, rather than small highly-correlated batches of long trajectories.
This further highlights the flexibility and advantage of a modular design.

\begin{figure}
    \centering
    \includegraphics[width=0.99\linewidth]{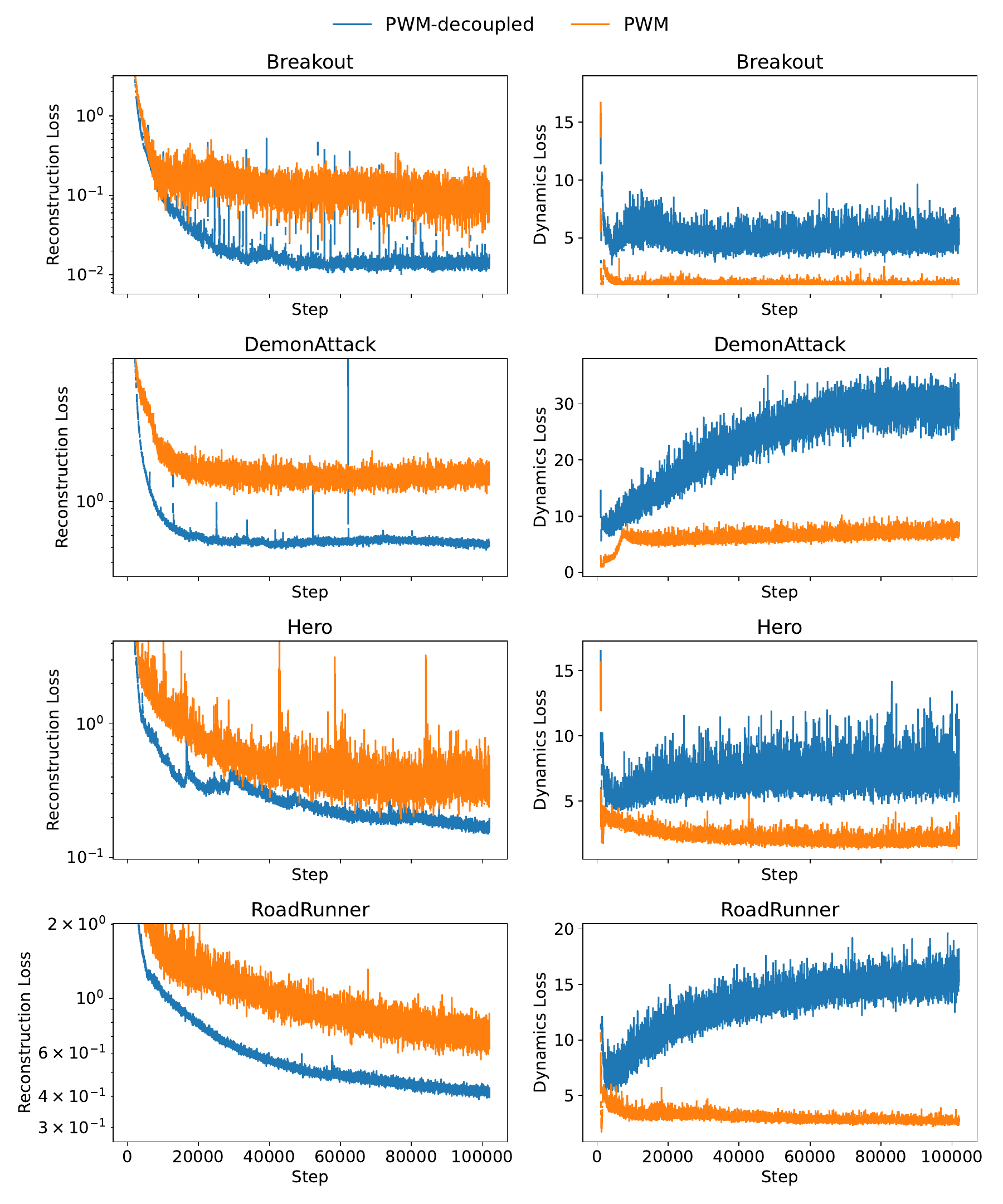}
    \caption{Reconstruction ($\mathcal{L}_{\text{pred}}$) and dynamics ($\mathcal{L}_{\text{dyn}}$) losses of PWM and PWM-decoupled on four Atari games (single seed). The first column uses a log-scaled y-axis. Decoupling the optimization objectives consistently reduces reconstruction loss while increasing dynamics loss, suggesting interference between the two objectives.}
    \label{fig:rssm-interference}
\end{figure}

\begin{figure}
    \centering
    \includegraphics[width=0.99\linewidth]{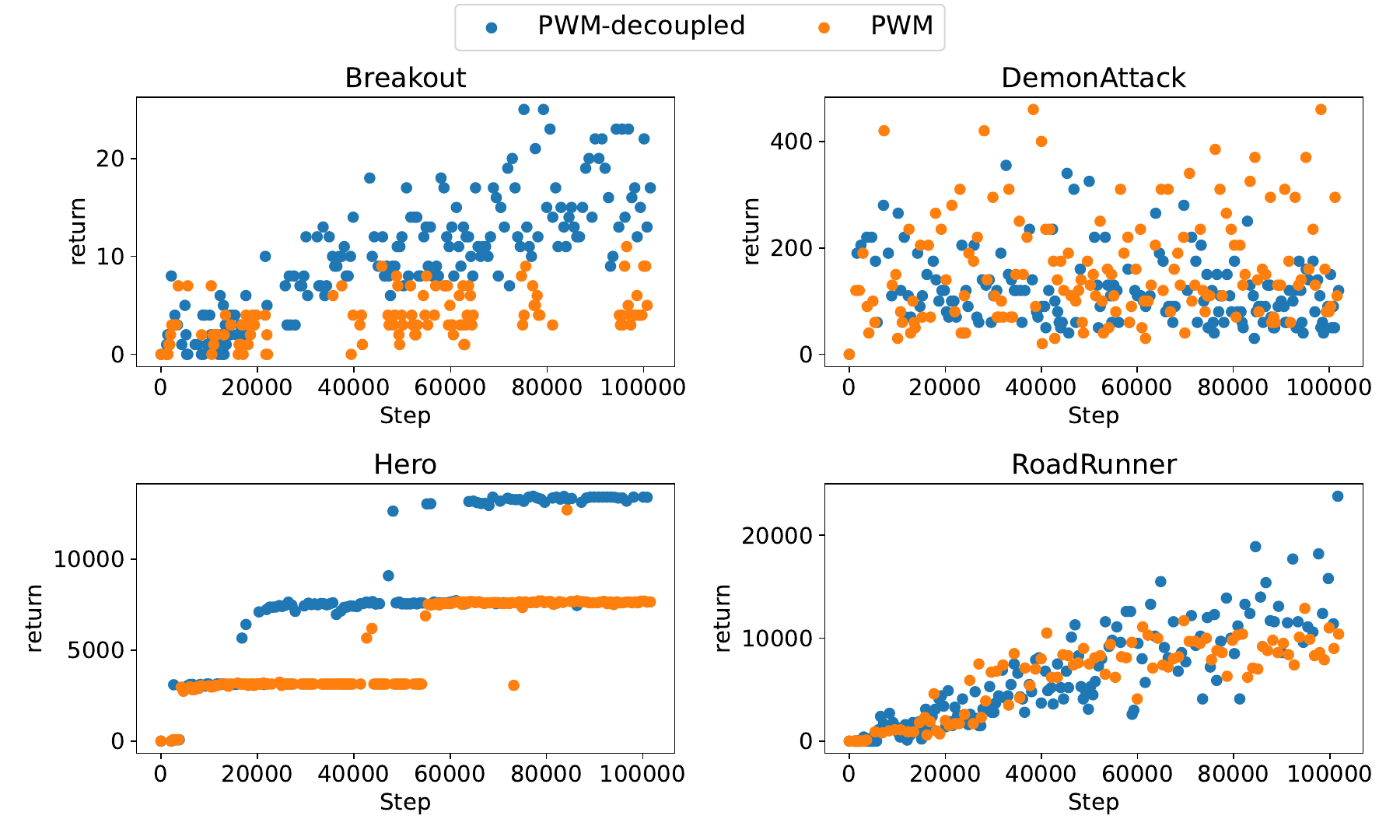}
    \caption{Agent episodic returns throughout training of PWM and PWM-decoupled on four Atari games (single seed). Each marker corresponds to a single episode.}
    \label{fig:rssm-interference-returns}
\end{figure}

\begin{figure}
    \centering
    \includegraphics[width=0.99\linewidth]{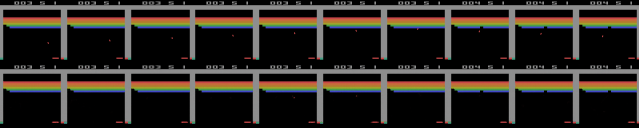}
    \caption{Ground truth (top) and reconstructed (bottom) frames from a training episode of PWM after 50K steps (half way though training). Notably, the ball is missing in most frames, suggesting the reason for its poor performance in this game.}
    \label{fig:rssm-interference-pwm-jux}
\end{figure}

\begin{figure}
    \centering
    \includegraphics[width=0.99\linewidth]{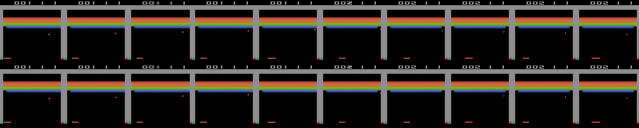}
    \caption{Ground truth (top) and reconstructed (bottom) frames from a training episode of PWM-decoupled after 50K steps (half way though training). Here, the ball is reconstructed in most frames, demonstrating the significantly improved representation performance.}
    \label{fig:rssm-interference-pwm-decoupled-jux}
\end{figure}

We compare the original PWM algorithm to its decoupled variant, PWM-decoupled, across four Atari environments: \texttt{Breakout}, \texttt{DemonAttack}, \texttt{Hero}, and \texttt{RoadRunner}.
These are games where PWM performed either particularly well (\texttt{Hero} and \texttt{RoadRunner}) or poorly (\texttt{Breakout} and \texttt{DemonAttack}).
Each variant is trained online from scratch on each game.
The results are presented in Figure~\ref{fig:rssm-interference}.
In addition, we present the achieved episodic returns in Figure~\ref{fig:rssm-interference-returns}, and the reconstruction quality of example episodes in Figure \ref{fig:rssm-interference-pwm-jux} (PWM) and Figure \ref{fig:rssm-interference-pwm-decoupled-jux} (PWM-decoupled).

Although our results are based on a single random seed and are limited to only four environments, we observe a consistent trend.
First, the reconstruction losses are consistently and significantly lower when decoupling the optimization, while the dynamics losses are significantly higher.
This suggests that the objectives are interfering.

Second, we observe similar or better episodic returns (Figure \ref{fig:rssm-interference-returns}) using the decoupled optimization, suggesting that the higher dynamics loss might not lead to worse world modeling performance in practice.
Note that a higher dynamics loss in this case does not necessarily mean worse performance, as for example multiple discrete combinations could represent the same or similar frame. Thus, when the dynamics model fails to predict a specific combination, it leads to high loss values while the underlying representations are accurate.

Lastly, we report that similar trends were observed when training only the world model in an offline, supervised-learning fashion on pre-collected datasets. We explored this setting to eliminate complexities that may arise due to the online collection of the data.

While the presented preliminary results are noisy and limited, we believe that they uncover an interesting observation on the design and optimization of current world models.

\newpage
\section{Implementation Details}
\label{sec:appendix-impl-details}

\paragraph{Ablation Studies}
For the Atari 100K benchmark, we conducted ablations on the following games: \texttt{Assault}, \texttt{Breakout}, \texttt{ChopperCommand}, \texttt{CrazyClimber}, \texttt{JamesBond}, \texttt{Kangaroo}, \texttt{Seaquest}, and \texttt{UpNDown}.
For the DeepMind Control Suite, we used the tasks: \texttt{acrobot-swingup}, \texttt{cartpole-swingup-sparse}, \texttt{cheetah-run}, \texttt{finger-turn-hard}, \texttt{hopper-stand}, \texttt{pendulum-swingup}, \texttt{reacher-hard}, and \texttt{walker-run}.

\paragraph{Code}
We \href{https://anonymous.4open.science/r/M3-3457/}{open-source our code and trained model weights}.
Our code is written in Pytorch \citep{Paszke2019Pytorch}.

\paragraph{Hardware} 
All Atari and DMC experiments were performed on V100 GPUs, while for Craftax a single RTX 4090 was used.

\paragraph{Run Times}
Experiments on Atari require approximately 12 hours on an RTX 4090 GPU and around 29 hours on a V100 GPU. For DMC, the runtime is about 40 hours on a V100 GPU. Craftax runs take roughly 94 hours, equivalent to 3.9 days.

\paragraph{Craftax}
The official environment provides the categorical variables in one-hot encoding format.
Our implementation translates these variables to integer values which can be interpreted as tokens.

\paragraph{Setup in Atari Freeway}
For the Freeway environment, we adopted a modified sampling strategy where a temperature of 0.01 is used instead of the standard value of 1, following \citep{micheli2022transformers, cohen2024improving}. 
This adjustment helps directing the agent toward rewarding paths. 
Note that alternative approaches in the literature tackle the exploration challenge through different mechanisms, including epsilon-greedy exploration schedules and deterministic action selection via argmax policies \citep{micheli2022transformers}.


\end{document}